\documentclass[3p,authoryear]{elsarticle}
\makeatletter
\def\ps@pprintTitle{%
  \let\@oddhead\@empty
  \let\@evenhead\@empty
  \def\@oddfoot{\reset@font\hfil\thepage\hfil}
  \let\@evenfoot\@oddfoot
}
\makeatother

\usepackage{graphicx}
\usepackage{subfig}
\usepackage{amsmath, amsfonts, amssymb, bm}

\usepackage{geometry}%
\usepackage{fourier}
\DeclareMathAlphabet{\mathcal}{OT1}{pzc}{m}{it}
\DeclareSymbolFont{letters}{OML}{cmm}{m}{it}

\graphicspath{{./Graphics/}}

\usepackage{color, xcolor, colortbl}
\usepackage{hyperref}
\hypersetup{
  colorlinks,
  citecolor=blue,
  filecolor=blue,
  linkcolor=blue,
  urlcolor=blue
}

\begin{document}
\begin{frontmatter}
  \title{Nested Cavity Classifier:\\performance and remedy\tnoteref{t1}} \tnotetext[t1]{This
    manuscript was initially composed in 2009 as part of a research pursued that time. This paper is
    currently under consideration in Pattern Recognition Letters.}  \author[WAM]{Waleed~A.~Mustafa}
  \ead{WaleedAMustafa@gmail.com}

  \author[WAY]{Waleed~A.~Yousef\corref{cor1}}
  \ead{wyousef@GWU.edu, wyousef@fci.helwan.edu.eg}
  \cortext[cor1]{Corresponding Author}

  \address[WAM]{B.Sc., TU Kaiserslautern: Technische Universit\"at Kaiserslautern}

  \address[WAY]{Ph.D., Associate Professor, Computer Science Department, Faculty of Computers and
    Information, Helwan University, Egypt.\\ Director, Human Computer Interaction Laboratory (HCI Lab.)}

  \begin{abstract}
    Nested Cavity Classifier (NCC) is a classification rule that pursues partitioning the feature
    space, in parallel coordinates, into convex hulls to build decision regions. It is claimed in
    some literatures that this geometric-based classifier is superior to many others, particularly
    in higher dimensions. First, we give an example on how NCC can be inefficient, then motivate a
    remedy by combining the NCC with the Linear Discriminant Analysis (LDA) classifier. We coin the
    term Nested Cavity Discriminant Analysis (NCDA) for the resulting classifier. Second, a
    simulation study is conducted to compare both, NCC and NCDA to another two basic classifiers,
    Linear and Quadratic Discriminant Analysis. NCC alone proves to be inferior to others, while
    NCDA always outperforms NCC and competes with LDA and QDA.
  \end{abstract}

  \begin{keyword}
    Classification \sep Nested Cavity Classifier (NCC) \sep Parallel Coordinates.
  \end{keyword}

\end{frontmatter}

\section{Introduction}\label{sec:Introduction}
Nested Cavity Classifier (NCC) is a geometric-based classification rule that pursues partitioning
the feature space in parallel coordinates (abbreviated $||$-coords) into convex-hulls to build
decision regions \citep{Inselberg2000ClassVisu}. Many articles and books considered the assessment
of classifiers using simulated and real-world datasets (e.g., \citep{Raudys1980DimenSampleSize,
  Efron1997ImprovementsOnCross, Hastie2001TheElements}); but none of them considered a systematic
assessment of NCC. However, \cite{Inselberg2000ClassVisu} compared NCC with other classifiers only
on few real high-dimensional datasets; that study mentioned the superiority of NCC over other
classifiers.

NCC, as described below, builds decision regions geometrically using convex hulls. This partitioning
mechanism has a drawback on the performance of the NCC (as explained in Section \ref{sec:ncc}). NCC
classifies any testing observation---regardless to its class, whether ``class 1'' or ``class
2''---as class, say, ``class 2'' as long as it does not lie inside the range of the training data
set; i.e., within the minimum and maximum values of each dimension. Since this is not always true,
the present article proposes combining NCC with LDA to classify observations outside the range of
the training set. We coin ``Nested Cavity Discriminant Analysis (NCDA)'' as a name for the resulting
classifier.

% \bigskip

The present article is organized as follows.
% Section \ref{SecParallelCoord} provides a review on parallel coordinates ($||$-coords, as abbreviated in \cite{Inselberg2000ClassVisu}).
Section \ref{sec:ncc} explains the NCC. Section \ref{sec:ncclda} motivates combining NCC with
LDA. Section \ref{SecSimulation} is a simulation study that compares NCC and NCDA to other
classifiers. Section \ref{sec:conclusion} is a conclusion and a discussion for future work.

\section{Parallel Coordinates ($||$-Coords)}\label{SecParallelCoord}
Data visualization can inspire one to solve very complex problems. When data is visualized,
inter-variable relations can be easily spotted; these relations are patterns.  Detecting these
patterns is a pattern recognition problem. We usually map problems into geometrical space; and by
using the amazing pattern recognition capabilities of our eyes and brains we try to figure things
out.

Mapping a problem into the ordinary geometrical space involves mapping variables into corresponding
space axes (orthogonal axes). A problem arises when there is a need to visualize high dimensional
data because we are only familiar with 3 dimensional orthogonal space. This confines us to visualize
only 3-dimensional problems, which is not sufficient in real-life situations. Said differently,
``\textit{orthogonal visualization uses up the plane very quickly}'' \cite{Inselberg2002VisData}.

Orthogonality, depending on the notion of an ``angle'', inspired ``Maurice d'Ocagne'' in 1885, who
realized that if we could represent the problem into axes without the need for an angle we will not
use orthogonal axes. This implies that we will not use up the plane that quickly. Since the opposite
of orthogonality is parallelism, representing the problem in a geometric form by mapping the
variables into parallel axes rather than orthogonal axes will help us to visualize high dimensional
problems.

\section{Nested Cavity Classifier (NCC)}\label{sec:ncc}
We first consider the binary classification problem, where an observation $t_i = (x_i, y_i)$ has the
$p$-dimensional feature vector $x_i$ (the predictor), and the response is $y_i$. The response $y_i$
equals one of the two classes, $\omega_1$ or $\omega_2$. Assume the availability of a training dataset
$\mathbf{t} = \lbrace t_i : t_i = (x_i, y_i),\ i = 1,\ldots, n \rbrace$. This data set is used to
learn the geometrical structure of the problem and to design a classification rule
$\eta_\mathbf{t}$. For any future observation having a feature vector $x_0$ and an unknown class,
the task of $\eta_\mathbf{t}$ is to predict the class $y_0$, i.e., to provide the prediction
$\eta_\mathbf{t}(x_0)$, which equals $\omega_1$ or $\omega_2$.

The process of learning from data and choosing a model for building $\eta_\mathbf{t}$ should be
preceded by data visualization, which provides a very useful insight to select the right model. NCC
works on the visualized version of the training data set represented in $||$-coords to build hyper
decision surfaces \citep{Inselberg2002VisData, Inselberg2000ClassVisu,
  Chen2008DataVisualization}. In the following paragraph we give a very brief account for
$||$-coords, then explain how NCC works.

\bigskip

\begin{figure}[t]\centering
  \subfloat[Polygonal line $\overline{C}$ represents the point
  $C = (c_1, c_2, c_3, c_4,
  c_5,c_6)$.]{\label{figure1}\includegraphics[width=0.45\textwidth,height=1.3in]{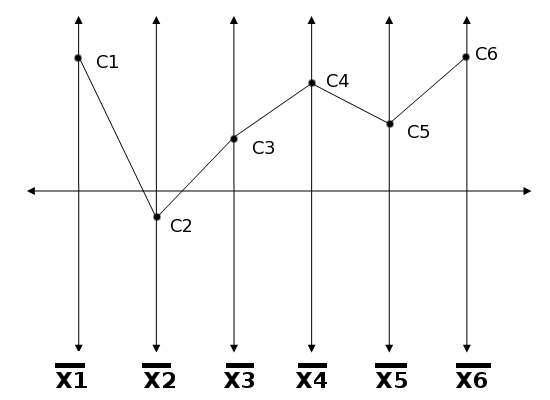}}
  \hspace{0.05\textwidth} \subfloat[Line interval in 9-D. Heavier polygonal lines represent
  endpoints.]{\label{figure2}\includegraphics[width=.45\textwidth,height=1.3in]{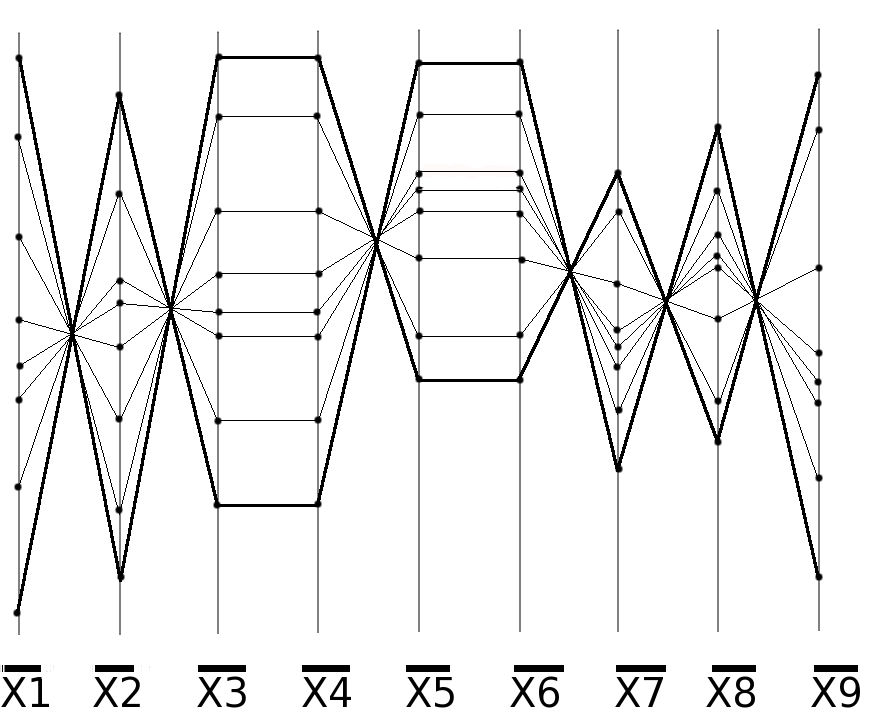}}
  \caption{Parallel Coordinates.}
  \label{FigParCord}
\end{figure}

In $X$-$Y$ Cartesian coordinates, $p$ copies of real lines labeled
$\overline{X}_1, \overline{X}_2, \ldots, \overline{X}_p$ are placed equidistantly and perpendicular to
the $X$-axis. These are the axes of the Parallel Coordinate system for $\mathcal{R}^p$ Euclidean
space. %All having the same positive orientation as the $Y$-axis.
A point $C$ with coordinates $(c_1, c_2, \dots, c_p)$ is represented by the complete polygonal line
$\overline{C}$, whose $p$ vertices are at $(i-1,c_i)$ on the $X_i$-axis, $i = 1,\dots, p$ as shown
in Figure \ref{FigParCord}-\subref*{figure1}. In this way, a 1-1 correspondence between points in
$\mathcal{R}^p$ and planar polygonal lines in $X$-$Y$ Cartesian coordinates is established
\cite{Inselberg2002VisData}. For Example, Figure \ref{FigParCord}-\subref*{figure2} shows another
example, a line segment in 9 dimensions by showing 8 points on this segment including the two
endpoints. This kind of data representation is impossible in the perpendicular coordinates.

\bigskip

\begin{figure}[t]\centering
  \subfloat[S11 and S12 represent S1.]{\label{fig:figure3}\includegraphics[width=0.28\textwidth]{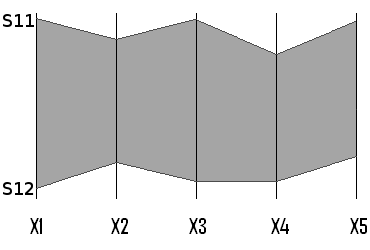}}\hspace{0.065\textwidth}
  \subfloat[S21 and S22 represent S2.]{\label{fig:figure4}\includegraphics[width=0.28\textwidth]{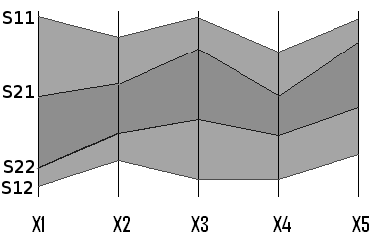}}\hspace{0.065\textwidth}
  \subfloat[S31 and S32 represent S3.]{\label{fig:figure5}\includegraphics[width=0.28\textwidth]{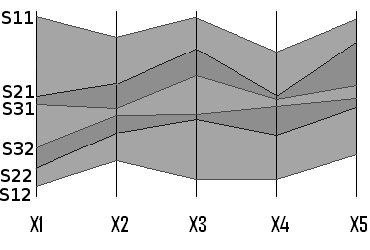}}
  \caption{Three regions built successively by NCC in $||$-coords.}\label{FigHypSur}
\end{figure}

In $||$-coords, a dataset with $p$ variables and $n$ observations is represented by a set of
2-dimensional points. The total number of those points equals $p \times n$. With the dataset
represented this way, we use an efficient convex-hull approximation algorithm to wrap (i.e., create
an approximate convex-hull) the points of, say, $\omega_1$. At this point we have created a hyper
surface $S_1$ that contains all observations of $\omega_1$ and some observations of $\omega_2$ as
well (Figure \ref{FigHypSur}-\subref*{fig:figure3}). We then apply convex-hull approximation to the
set of points of $\omega_2$ that are enclosed within the hyper surface $S_1$ to produce the hyper
surface $S_2$ (Figure \ref{FigHypSur}-\subref*{fig:figure4}). We repeat this process successively
till the maximum complexity required is reached (i.e. the maximum number of inner convex-hulls);
this is usually used as a regularization parameter to guard against overtraining.

After the algorithm terminates, the description of the hyper surface that represents the decision
region of $\omega_1$ is formalized as
\begin{equation}
  S_{\omega_1}= \left\{%
    \begin{array}[c]{lll}
      (S_1 - S_2) \cup (S_3 - S_4) \ldots \cup (S_{m-1} - S_{m}) &\text{if} & m\ \text{is even}\\
      (S_1 - S_2) \cup (S_3 - S_4) \ldots \cup S_m &\text{if} & m\ \text{is odd},
    \end{array}\right.
\end{equation}
where $S_m$ is the last produced hyper surface. Hence, for a given future observation $x_0$, the
classification rule $\eta_\mathbf{t}^{(NCC)}$ is formalized as follows.
\begin{equation}
  \eta_\mathbf{t}^{NCC}(x_0)  = \left\{%
    \begin{array}[c]{rl}
      \omega_1   &\text{if }   x_0  \in S_{\omega_1}\\
      \omega_2   &\text{otherwise}
    \end{array}
  \right.\label{EqNCCeta}
\end{equation}

\section{Combining LDA with NCC: a remedy}\label{sec:ncclda}
As described in section \ref{sec:ncc}, NCC uses a geometric criterion to build the decision
regions. The rule $\eta_\mathbf{t}^{(NCC)}$ in (\ref{EqNCCeta}) implies that
\begin{equation}
  \eta_\mathbf{t}^{(NCC)}(x_0) =\omega_2\,\ \forall x_0 \notin S_1.
\end{equation}
This means that all test observations belonging to $\omega_1$ and lying outside $S_1$ will be
misclassified!%---We may conjecture that the performance of this classification rule will deteriorate as $1/n$ increases faster than other rules. We will assess this conjecture in the simulation section.

\begin{figure}[t]\centering
  \includegraphics[height=1.5in,width=4in]{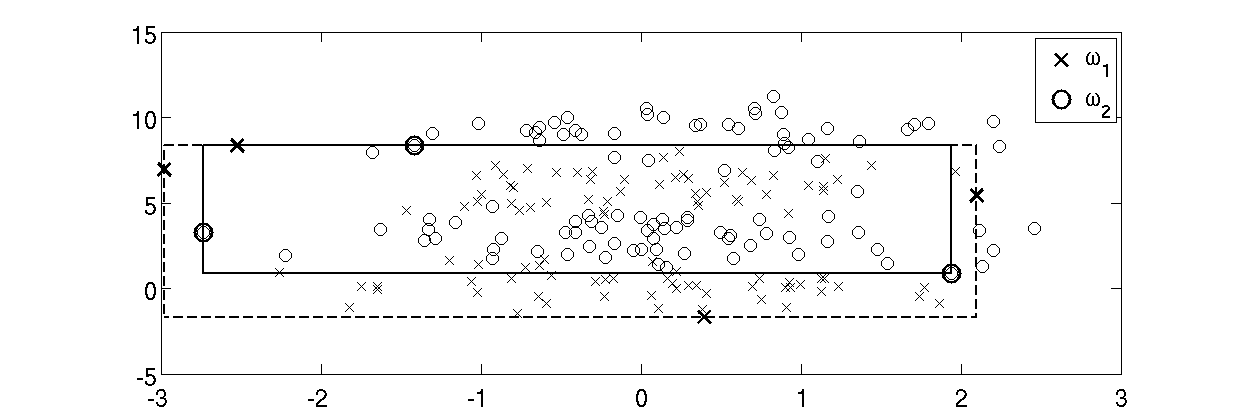}
  \caption{Decision regions built by NCC in $||$-coords and displayed in orthogonal coordinates. The
    dashed boundary represents $S_1$ and the solid represents $S_2$. The bold observations are those
    used to build the boundary.}\label{fig:parallelNCC}
\end{figure}

Figure \ref{fig:parallelNCC} illustrates decision regions built by NCC. The dashed rectangle
represents $S_1$ and the solid rectangle represents $S_2$. The observations used for building these
surfaces are plotted in bold. Notice that any observation $x_0\ (=(x_{01}, x_{02}))$ located outside
the boundaries of $S_1$ will be classified as $\eta_\mathbf{t}^{(NCC)}(x_0) = \omega_2$. However, from
the data plot, $x_0$ is most probably belonging to $\omega_1$ if $x_{02}<-2.2$ (the lower boundary of
$S_1$).

This was the motivation behind combining NCC with any other statistical rule. Such a combination
will provide the means for learning how to classify future observations located outside the outer
surface $S_1$. We chose the Linear Discriminant Analysis (LDA) for demonstration; however we could
have chosen the Quadratic Discriminant Analysis (QDA) or any other discriminant function, hence the
name Nested Cavity Discriminant Analysis (NCDA). However, since the aim behind this combination is
to be able to classify observations from the tail of the data distribution, we think that LDA or QDA
will be quite sufficient.

The proposed classification rule is simple; we train NCC using the data set $\mathbf{t}$ to learn
the geometric structure and build the decision rule $\eta_{\mathbf{t}}^{(NCC)}$. LDA is also trained
with $\mathbf{t}$ to build the decision rule $\eta_{\mathbf{t}}^{(LDA)}$. The final classification
rule $\eta_{\mathbf{t}}^{(NCDA)}$ will be
\begin{equation}
  \eta_{\mathbf{t}}^{(NCDA)}(x_0) = \left\{%
    \begin{array}{rl}
      \eta_{\mathbf{t}}^{(NCC)}(x_0) &\text{if } x_0 \in S_1 \\
      \eta_{\mathbf{t}}^{(LDA)}(x_0) &\text{otherwise,}
    \end{array}\right.
\end{equation}

\section{Simulation and Discussion}\label{SecSimulation}
A simulation study is conducted to compare NCC, NCDA, LDA, and QDA. several simulation parameters
should be considered; however, the purpose of the present article is to provide a preliminary
simulation study rather than a comprehensive one---refer to Section \ref{sec:conclusion} for future
work currently in progress. The data distributions $F_1$ and $F_2$ are assumed to be normals and
mixture of normals (with parameters discussed below), dimensionality $p$ is chosen to be 2, 4, 8 and
16, and size of the training set, $n$, is chosen to be 10, 20, 40, 80, 160 and 200 (assuming equal
training set sizes for the two classes).

We conducted three sets of experiments; in each experiment we train the classifier on a finite
training set (of the selected size $n$), and test on a testing set of size 1000 observations per
class to mimic the population. Each training and testing represents one Monte-Carlo (MC) trial. We
typically use 1000 MC trials; in each we train on a different training set (of the same size) drawn
from the same population and test on the same 1000-observation-per-class testing set.

We measure the performance in terms of error rate $Err$. The population parameters of interest,
then, are the mean (over the training sets of the same size) performance
$\operatorname*{E}_{\mathbf{tr}} Err$ and the variance $\operatorname*{Var}_{\mathbf{tr}}Err$. For
each experiment we plot the mean and the standard deviation of the performance versus the reciprocal
of the training set size.

\bigskip

\begin{figure}[htb]
  \centering
  \subfloat[$p=2$]
  {\label{fig:exp22dmean}\includegraphics[width=0.4\textwidth]{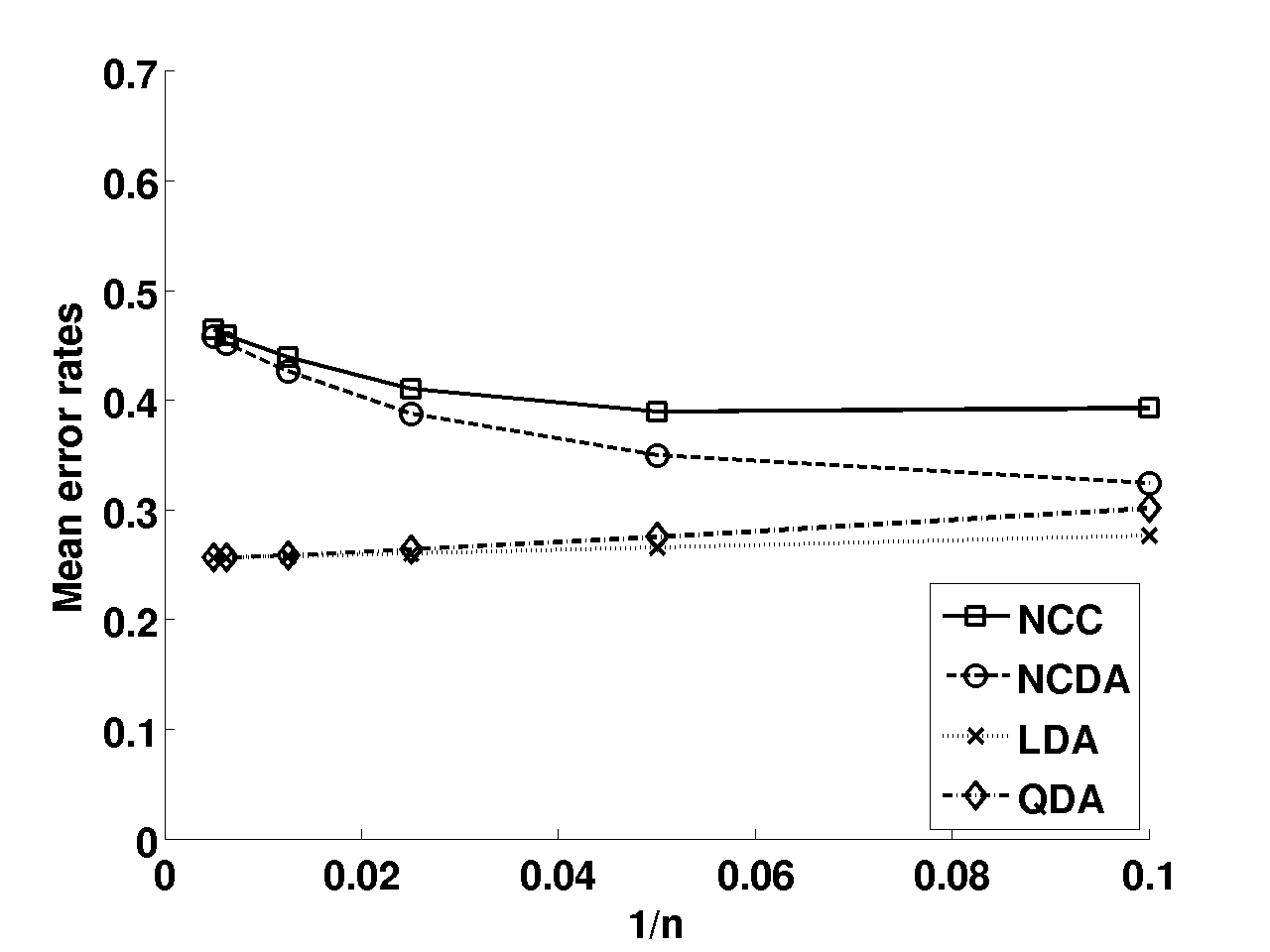}}
  \hspace{0.04\textwidth}
  \subfloat[$p=2$]
  {\label{fig:exp22dsigma}\includegraphics[width=0.4\textwidth]{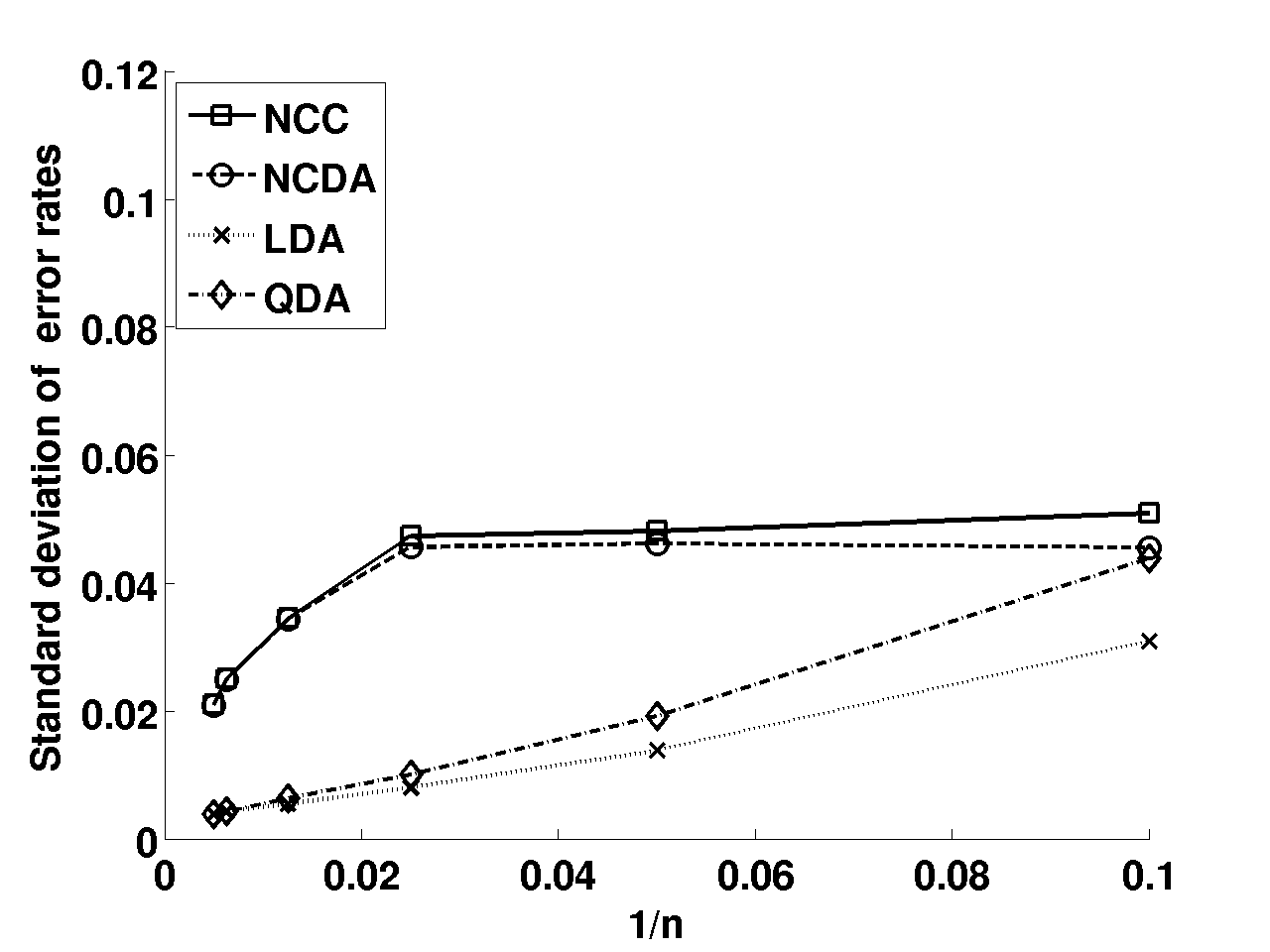}}
  \hspace{0.04\textwidth}
  \subfloat[$p=4$]
  {\label{fig:exp24dmean}\includegraphics[width=0.4\textwidth]{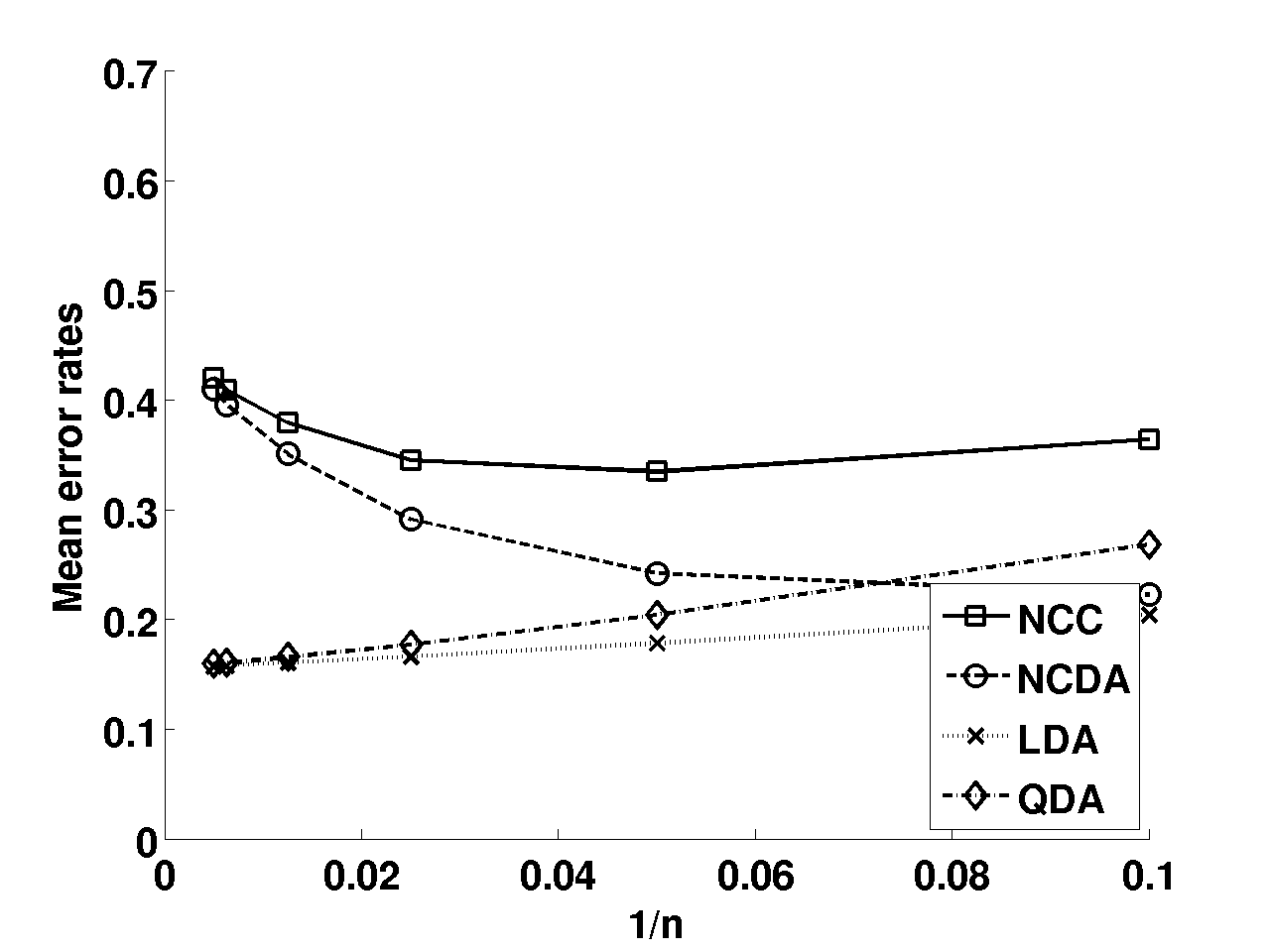}}
  \hspace{0.04\textwidth}
  \subfloat[$p=4$]
  {\label{fig:exp24dsigma}\includegraphics[width=0.4\textwidth]{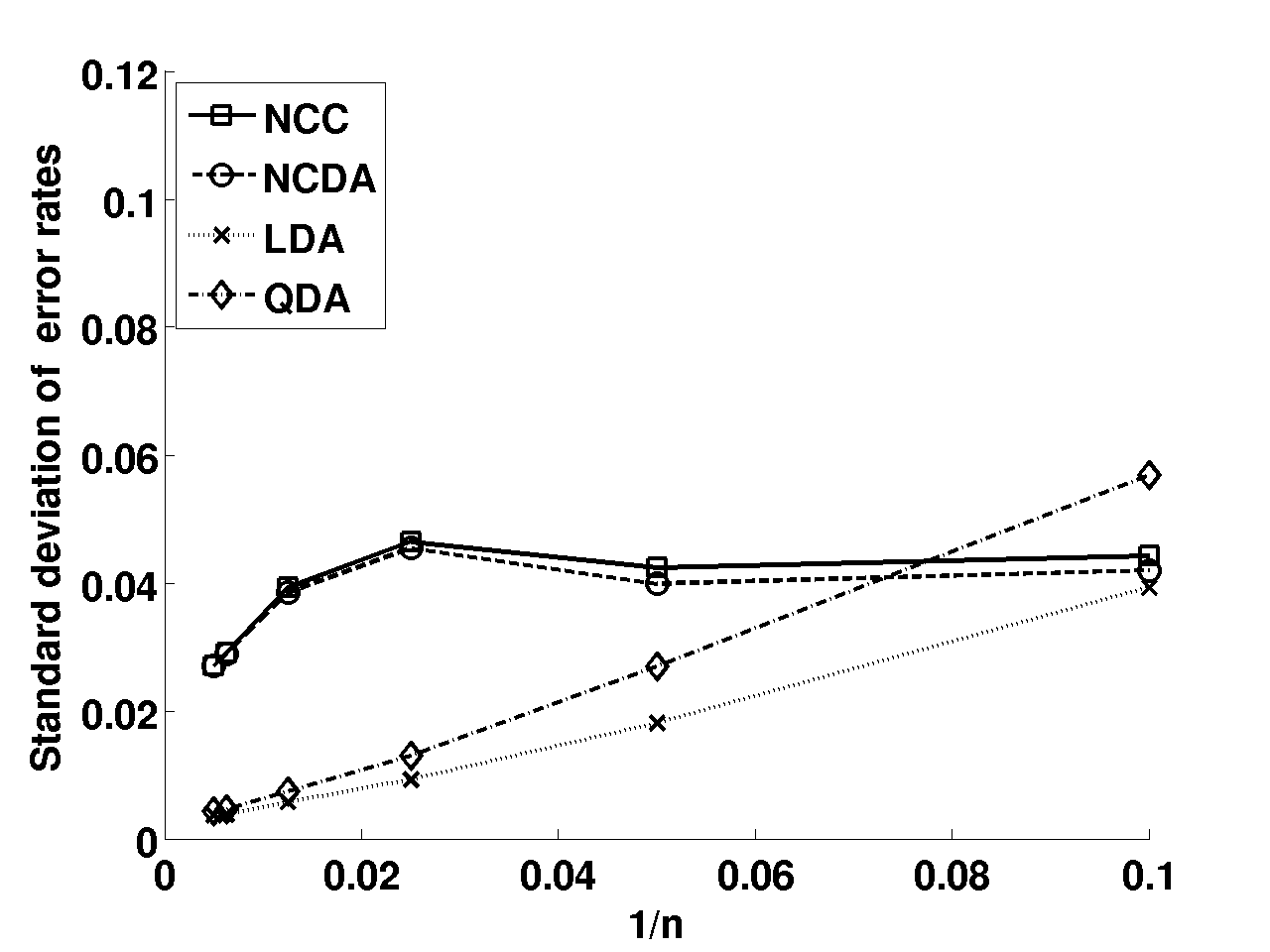}}
  \hspace{0.04\textwidth}
  \subfloat[$p=8$]
  {\label{fig:exp28dmean}\includegraphics[width=0.4\textwidth]{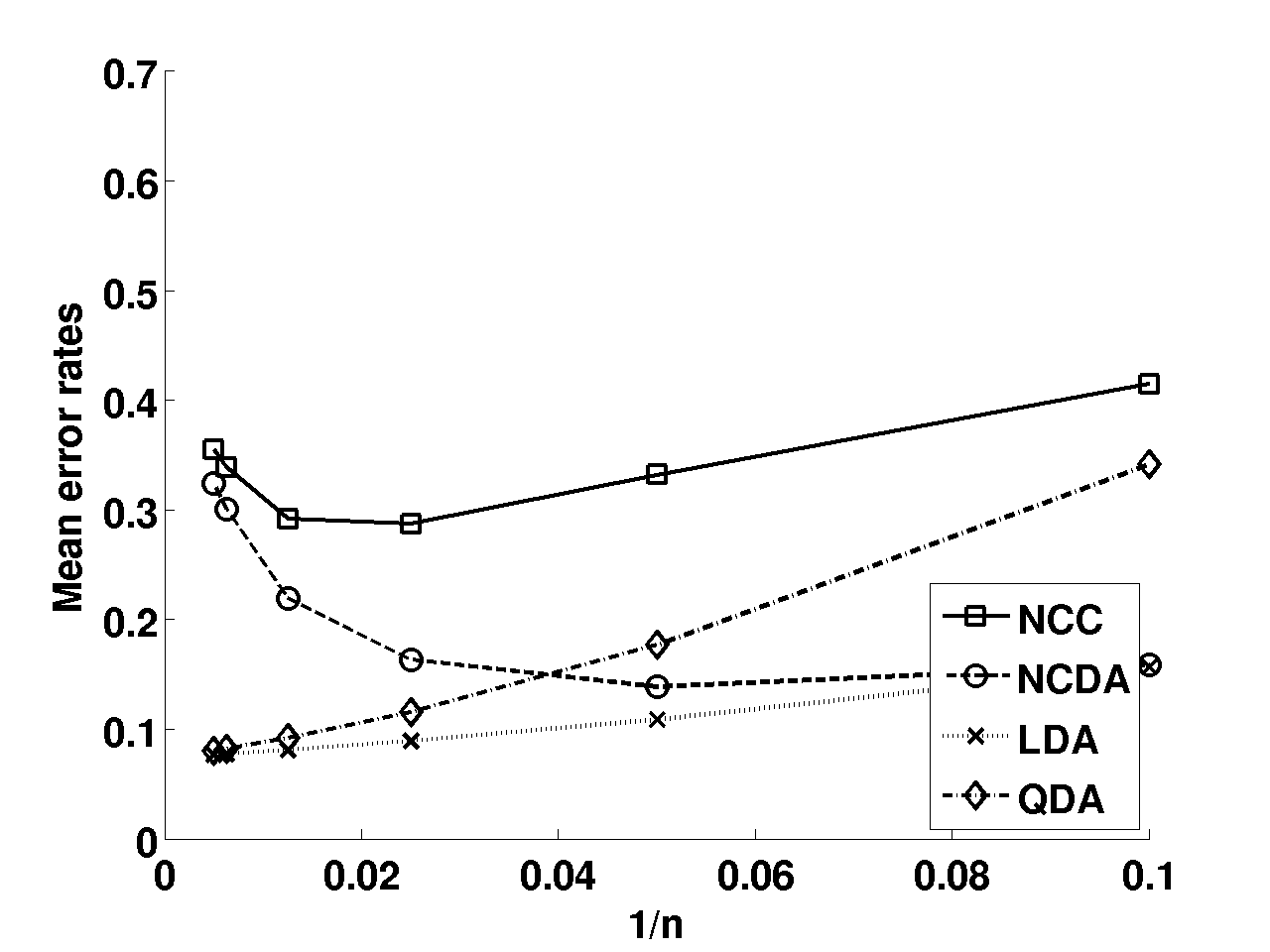}}
  \hspace{0.04\textwidth}
  \subfloat[$p=8$]
  {\label{fig:exp28dsigma}\includegraphics[width=0.4\textwidth]{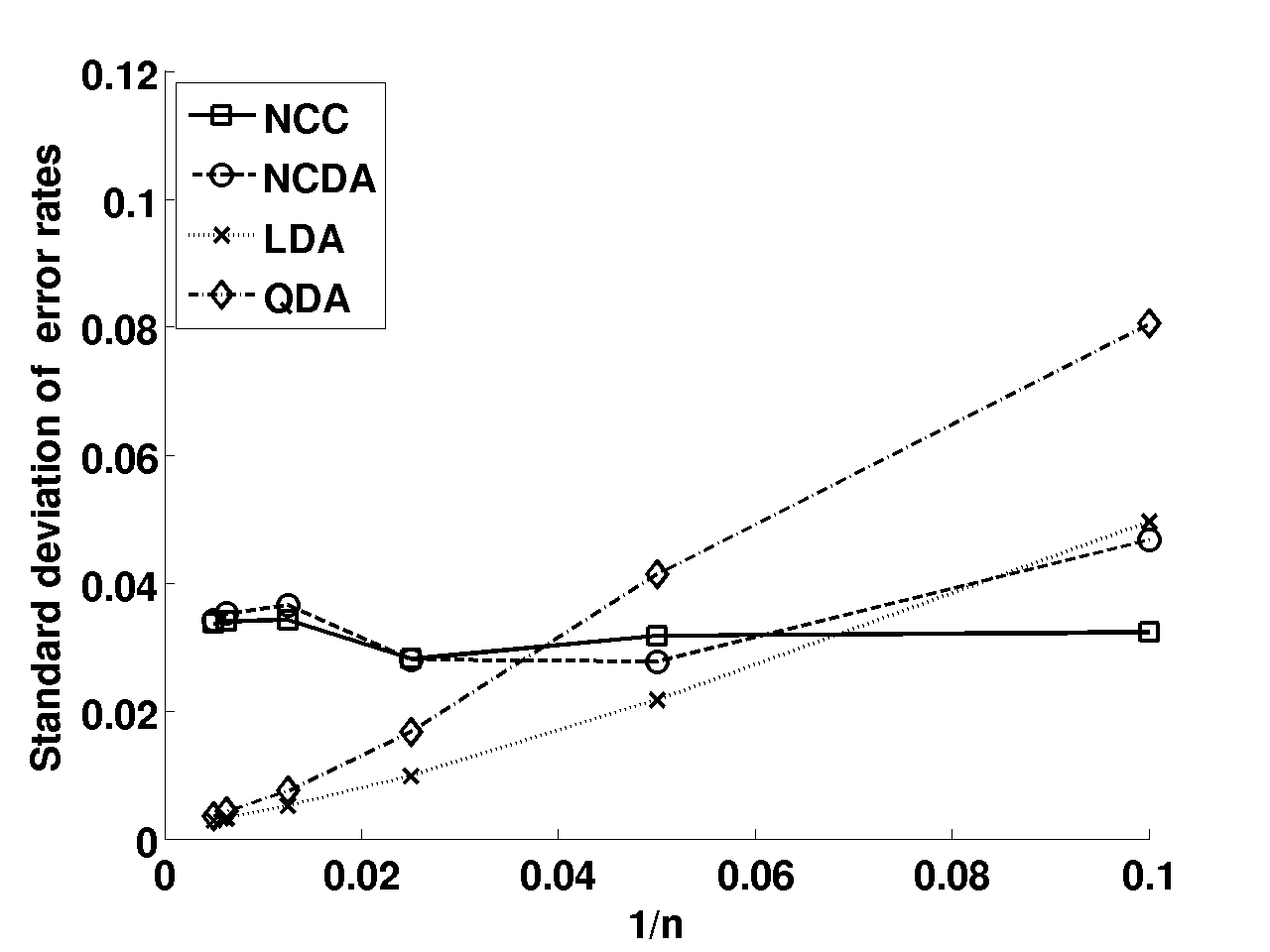}}
  \hspace{0.04\textwidth}
  \subfloat[$p=16$]
  {\label{fig:exp216dmu}\includegraphics[width=0.4\textwidth]{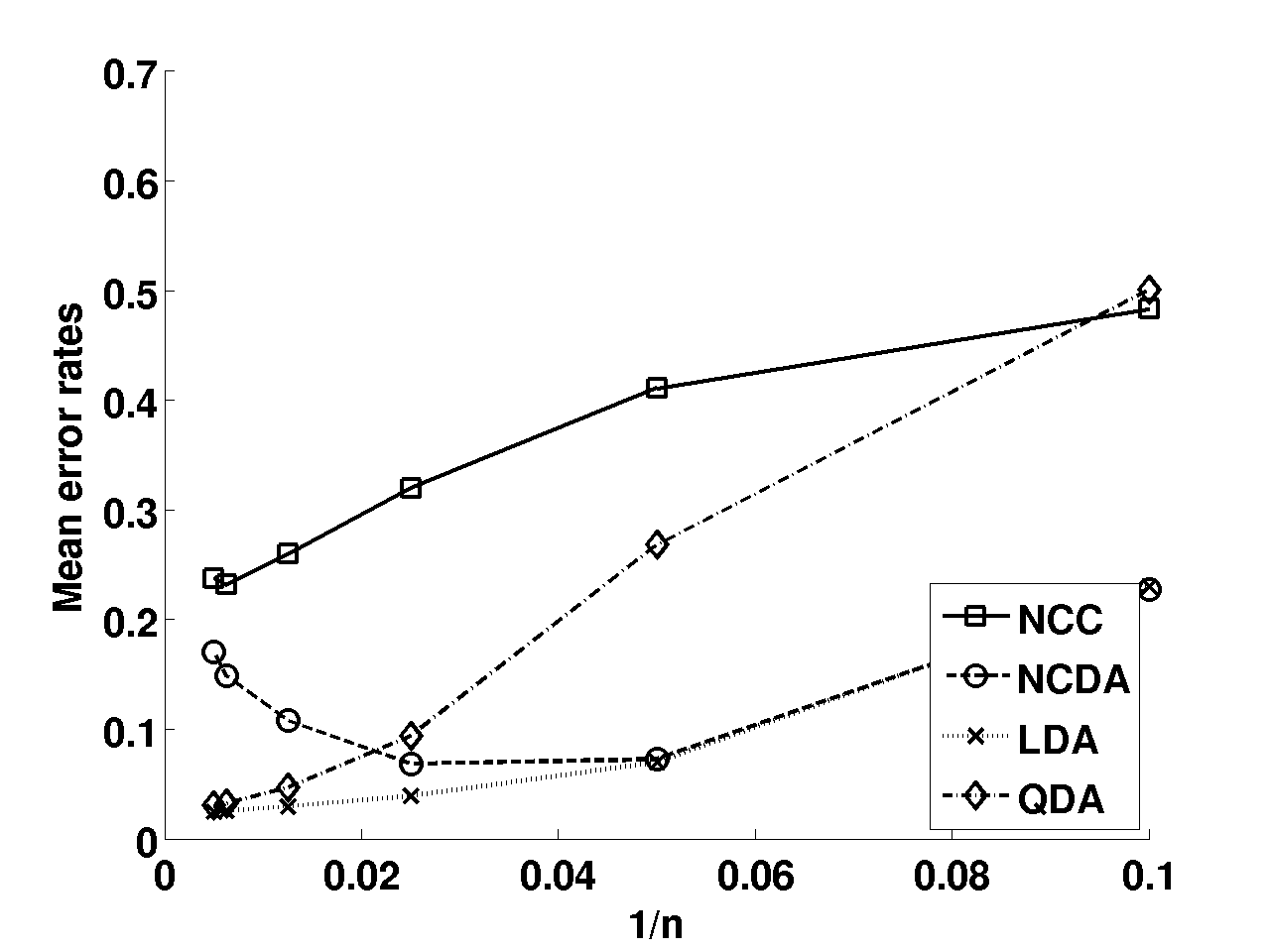}}
  \hspace{0.04\textwidth}
  \subfloat[$p=16$]
  {\label{fig:exp216dsigma}\includegraphics[width=0.4\textwidth]{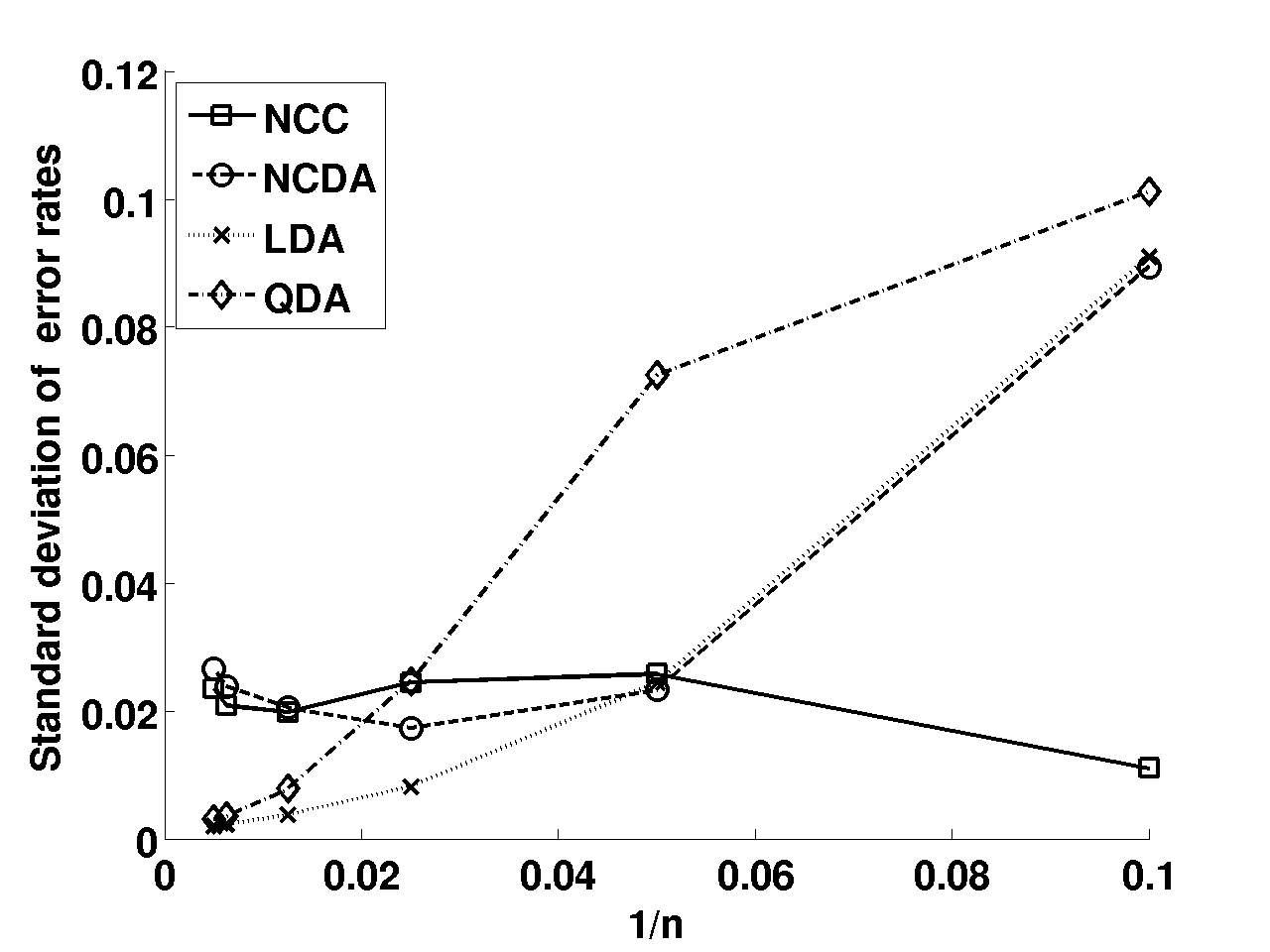}}
  \hspace{0.04\textwidth}
  \caption{Experiment 1 (multinormal distribution ): Mean (left) and standard deviation (right) of
    error rate under different $p$.}\label{FigExpNormal}
\end{figure}

The first set of experiments assumes $F_1$ and $F_2$ to be multinormal. The mean vector $\mu_1$ is set
to zero vector and the mean vector $\mu_2$ is set to $c\mathbf{1}$, where $\mathbf{1}$ is a vector
whose all components are equal to 1 and $c$ is a constant that can be used to adjust the classes
separability; for our current simulation study we set it to 1. Covariance matrices are set to
identity matrix. Figure \ref{FigExpNormal} presents the results of this configuration.

The figure illustrates the typical performance of the LDA and QDA that is well known in the
literature; \citep[e.g., see][]{Chan1998EffectsSampleSizeCAD}. The LDA is the winner if compared to
the other three, since it is the Bayes classifier for this configuration. However, the NCC behaves
the other way around for low dimensionality; its performance deteriorates as the training set size
increases! The interpretation of these results is interesting. For simplicity, consider the case of
$p=2$; the two distributions will look like two circular clouds, one centered at the origin and the
other is centered at $(c, c)$. With very small training data set, we can imagine that the mean
decision surface is a square centered at the origin and enclosed within the first cloud. This will
lead to a misclassification for all the testing observations coming from $F_1$ and occur outside the
square. Increasing the training set size gives a chance to more training observations to occur at
the tail of the distribution. Hence, this will widen the decision surface (the square)---refer to
Section \ref{sec:ncc} for information on how NCC works; this decreases the misclassification rate
from the first distribution. Therefore, the performance will increase with the training sample size
until the decision surface encloses---roughly speaking---the first cloud, yet, did not intersect
with the second cloud. Increasing the training sample size more will allow the decision surface to
grow until it intersects with the second cloud, the time at which the second type of error will
increase.

The improvement of NCDA over the NCC is evident for all dimensions and for all sample sizes; even,
it comes closer to the LDA for higher dimensions.

\bigskip

\begin{figure}[htb] \centering \subfloat[$p=2$]
{\label{fig:exp32dmean}\includegraphics[width=0.4\textwidth]{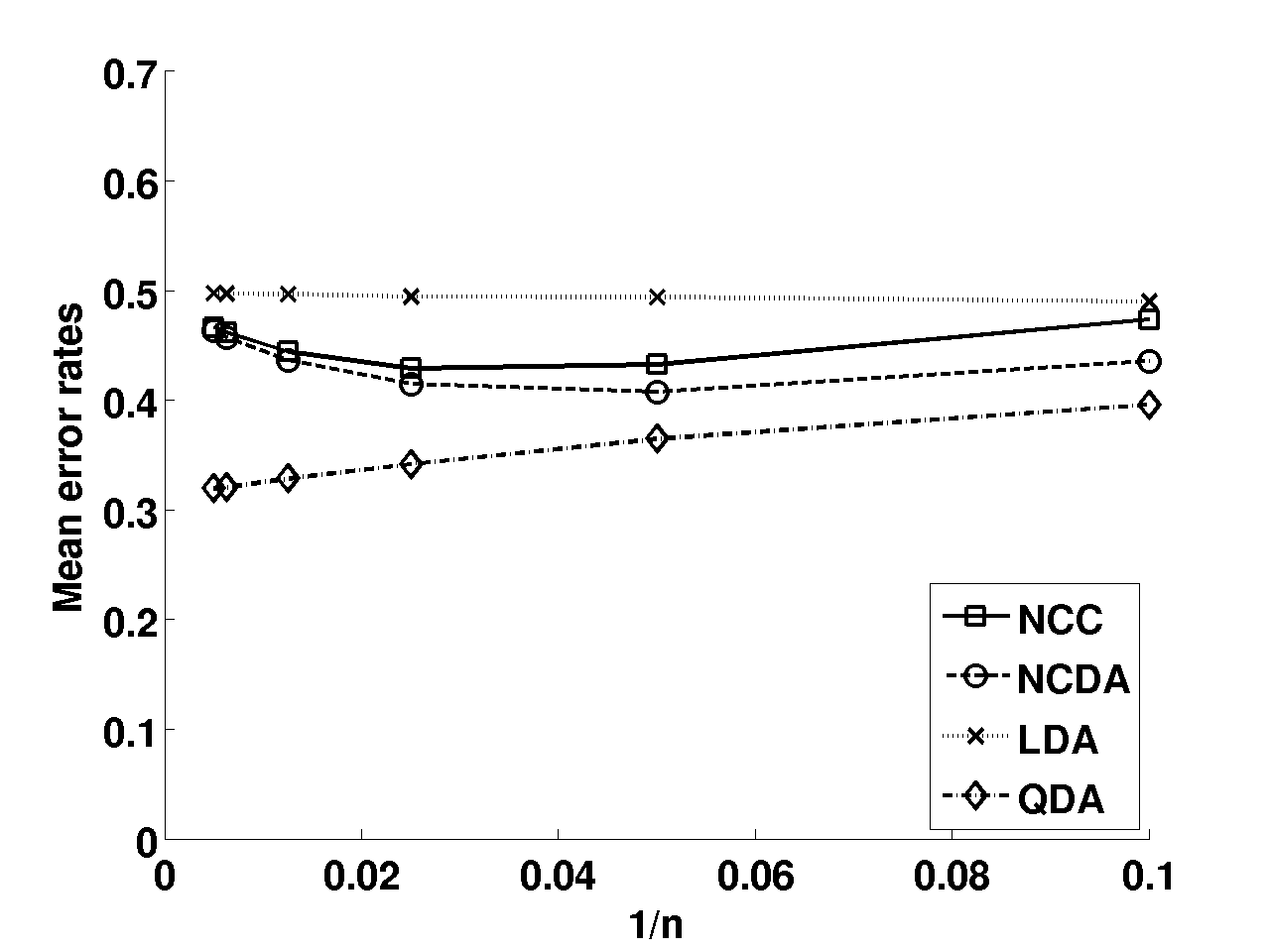}}
\hspace{0.04\textwidth} \subfloat[$p=2$]
{\label{fig:exp32dsigma}\includegraphics[width=0.4\textwidth]{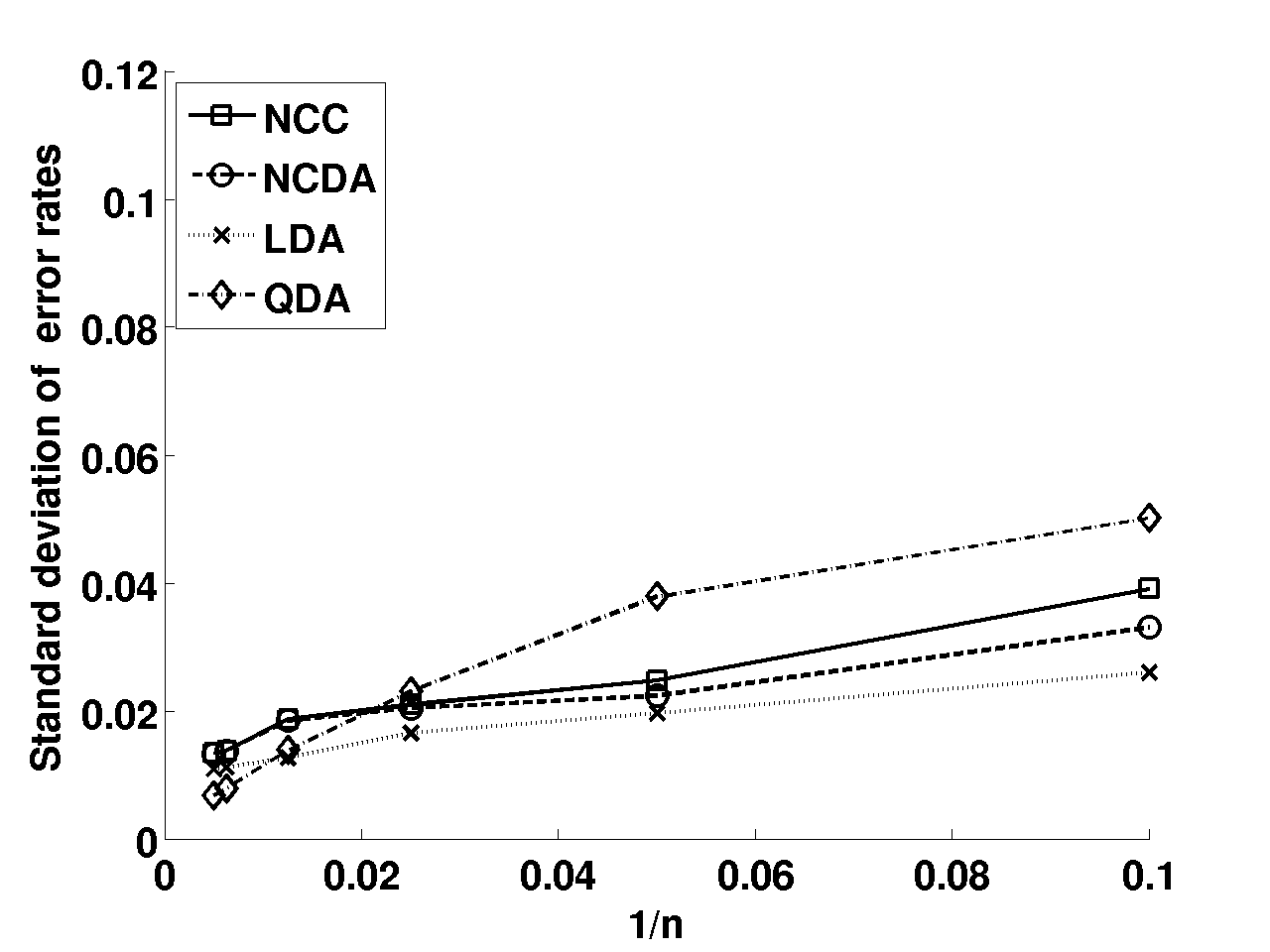}}
\hspace{0.04\textwidth} \subfloat[$p=4$]
{\label{fig:exp34dmean}\includegraphics[width=0.4\textwidth]{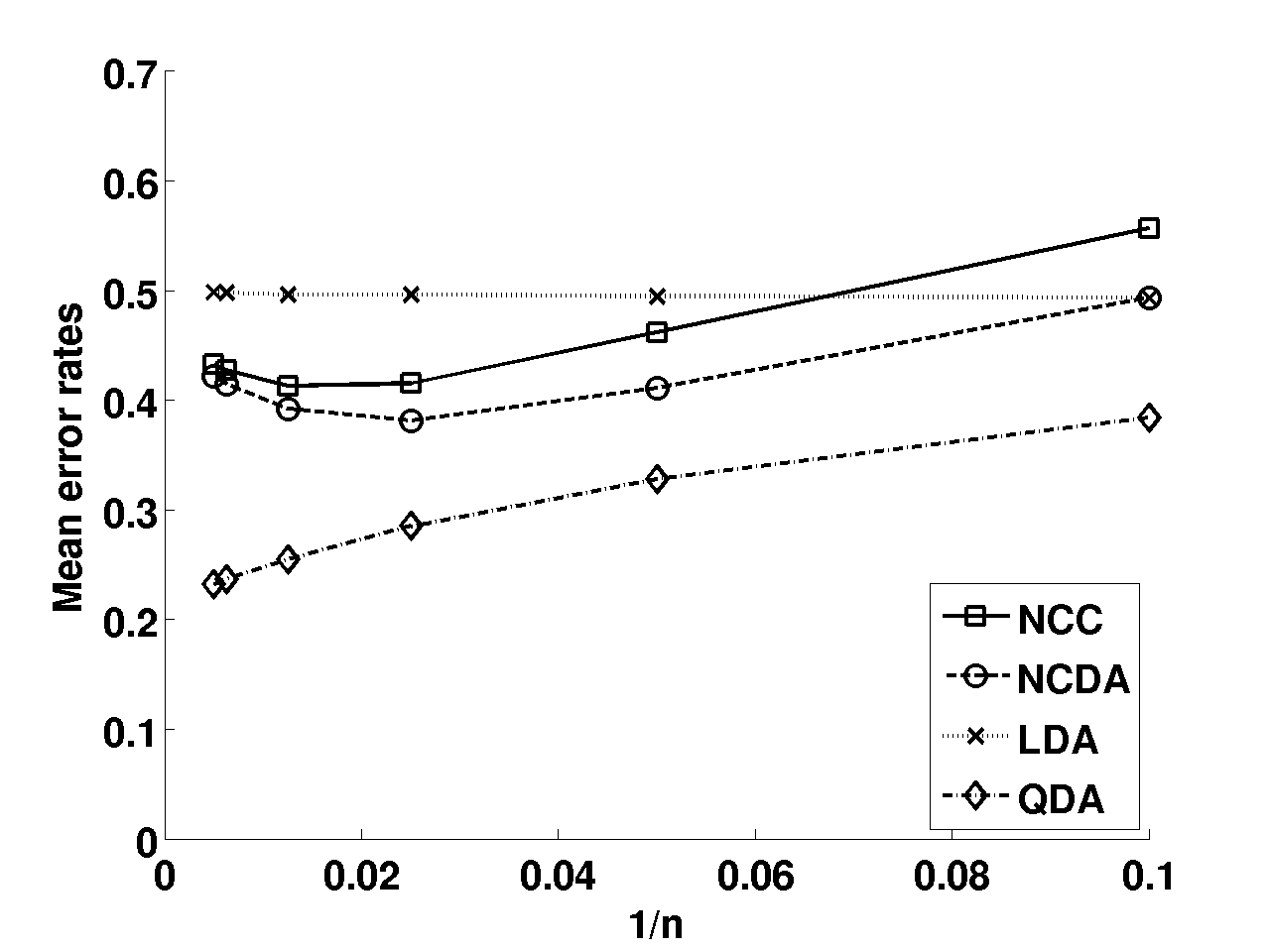}}
\hspace{0.04\textwidth} \subfloat[$p=4$]
{\label{fig:exp34dsigma}\includegraphics[width=0.4\textwidth]{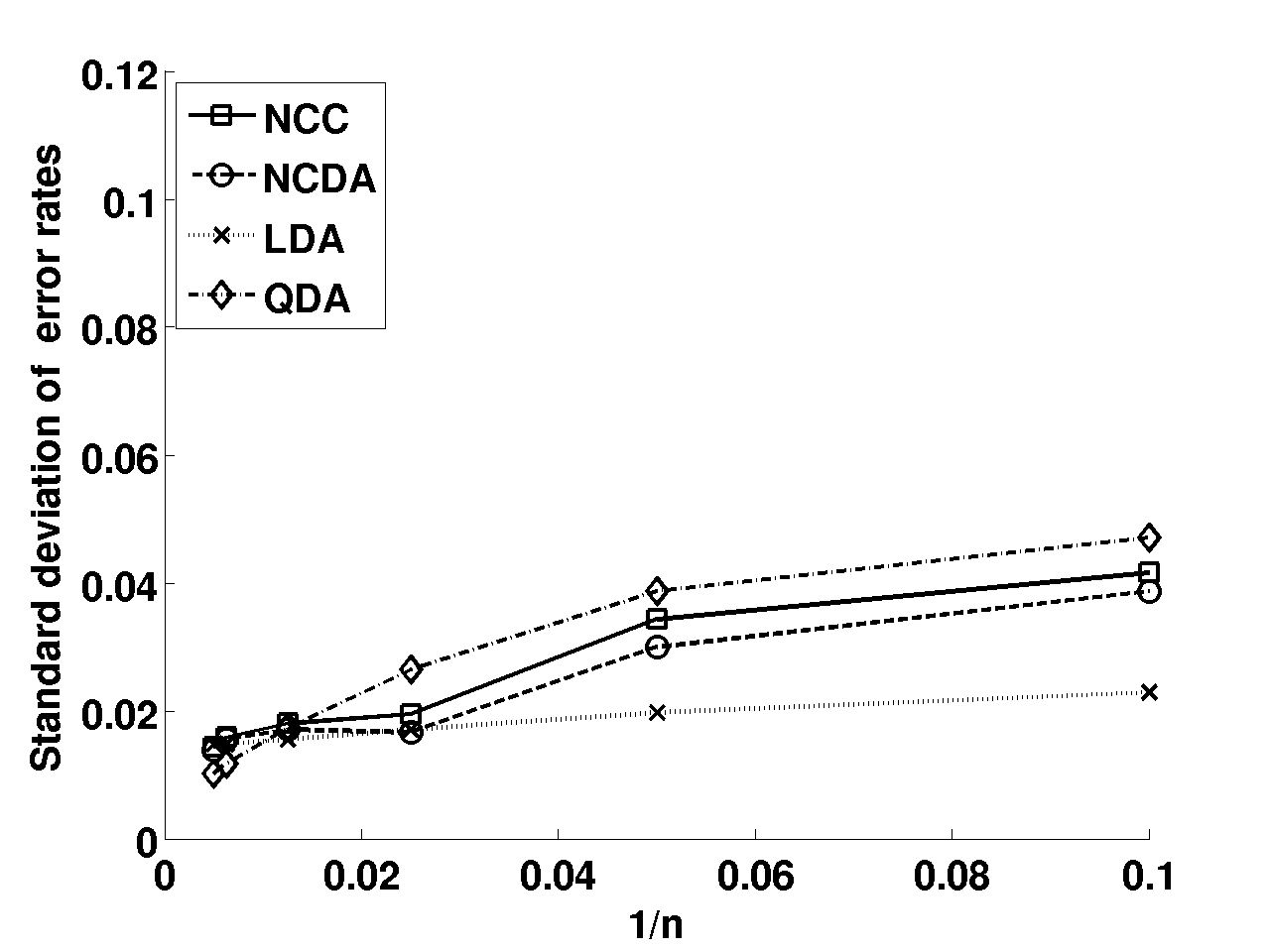}}
\hspace{0.04\textwidth} \subfloat[$p=8$]
{\label{fig:exp38dmean}\includegraphics[width=0.4\textwidth]{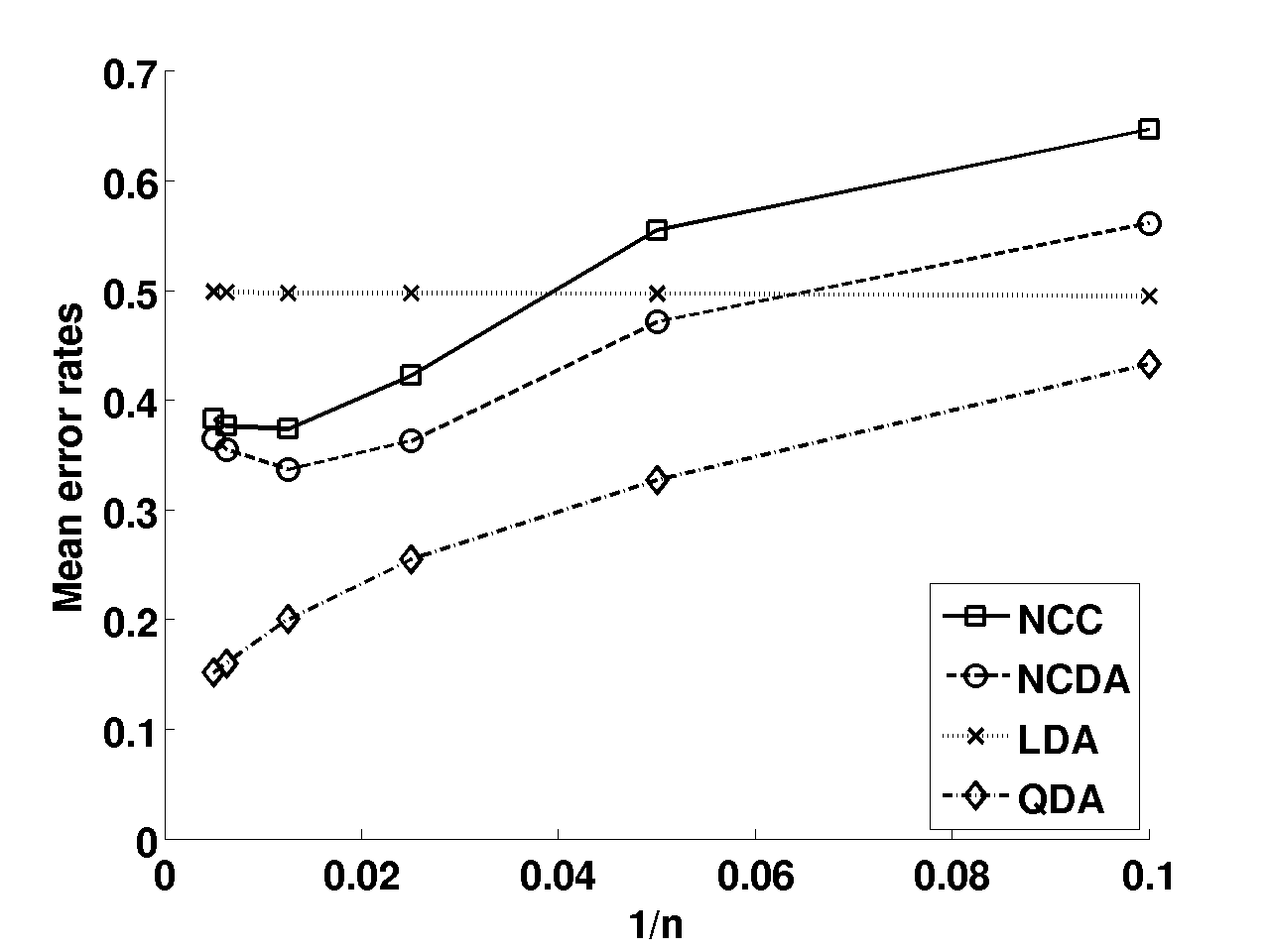}}
\hspace{0.04\textwidth} \subfloat[$p=8$]
{\label{fig:exp38dsigma}\includegraphics[width=0.4\textwidth]{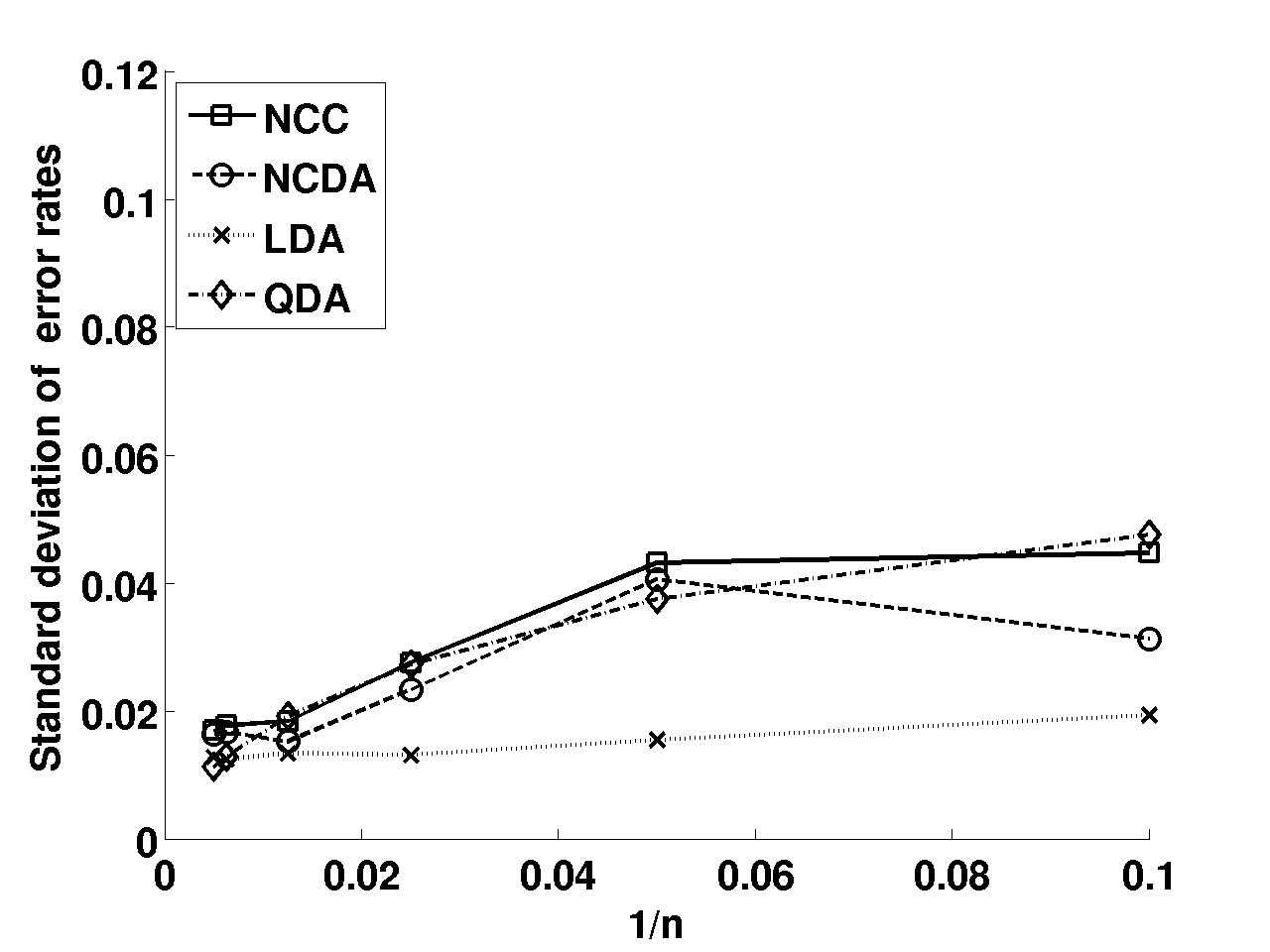}}
\hspace{0.04\textwidth} \subfloat[$p=16$]
{\label{fig:exp316dmu}\includegraphics[width=0.4\textwidth]{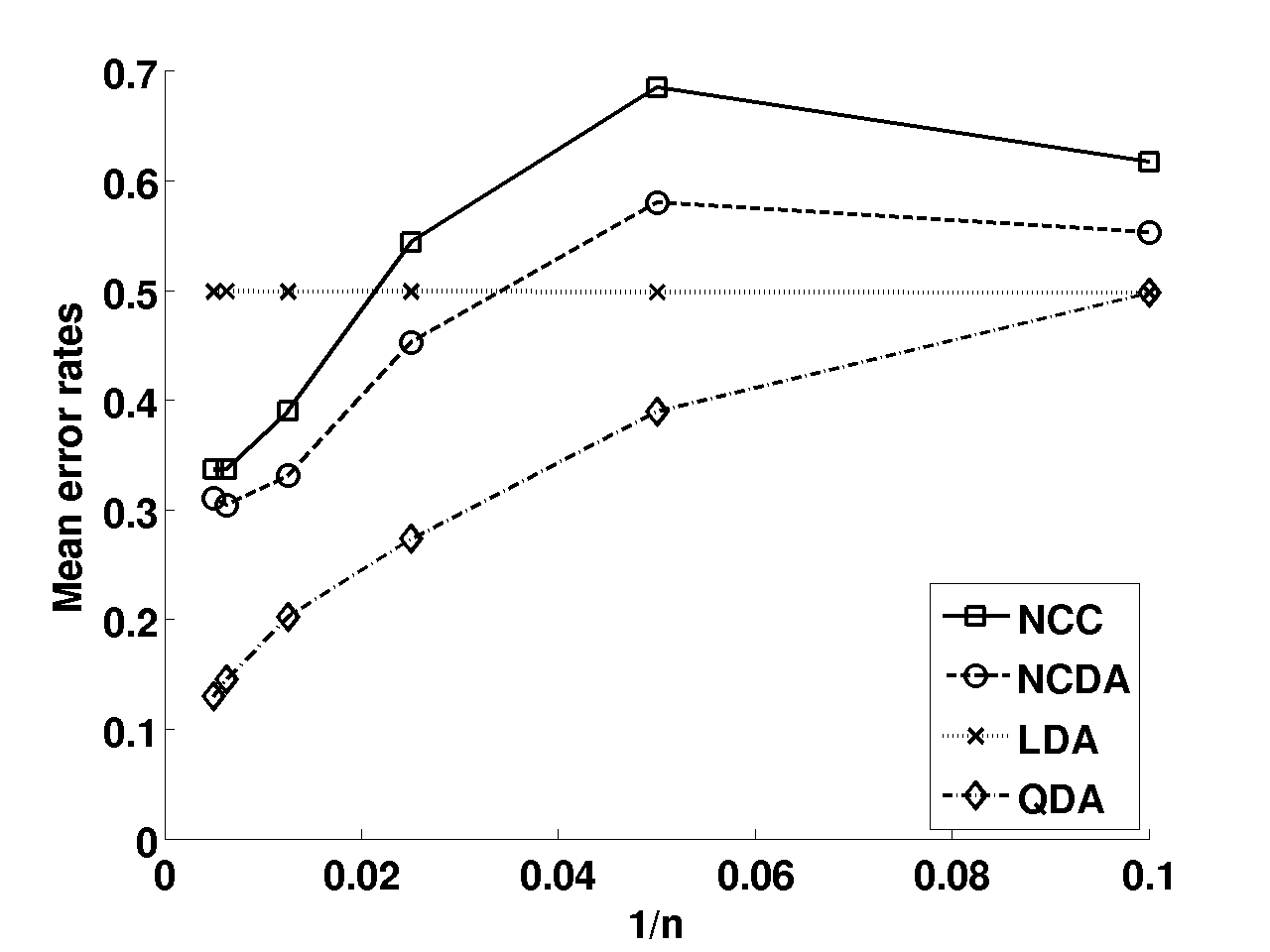}}
\hspace{0.04\textwidth} \subfloat[$p=16$]
{\label{fig:exp316dsigma}\includegraphics[width=0.4\textwidth]{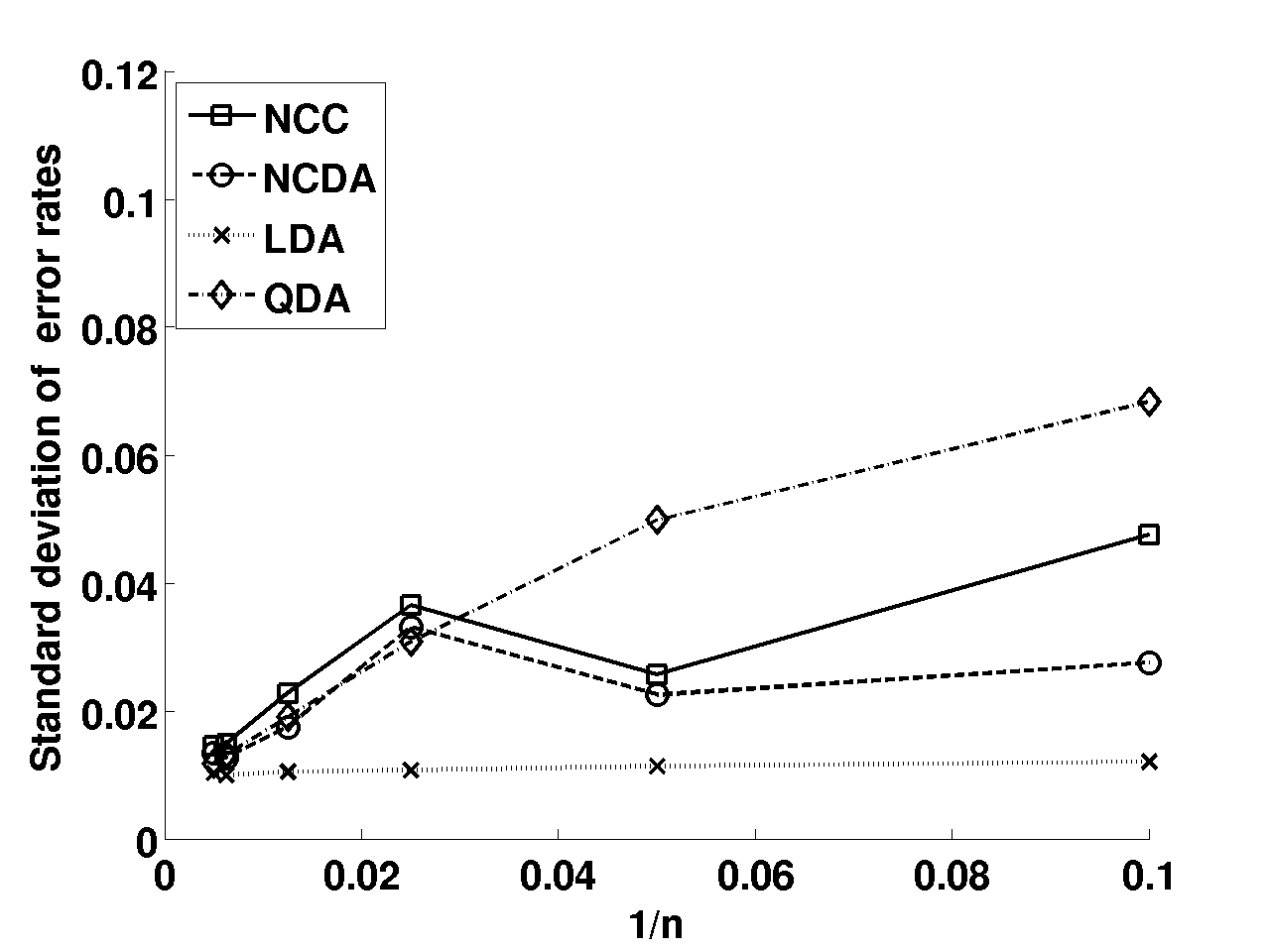}}
\hspace{0.04\textwidth}
\caption{Experiment 2 (mixture of two Gaussians): Mean (left) and standard deviation (right) of
  error rate under different $p$.}\label{FigExpNormMix}
\end{figure}

The second set of experiments assumes $F_1$ to be a mixture of two Gaussians, named $F_{11}$ and
$F_{12}$, with identity covariance matrices and mean vectors $\mu_{11} = \mathbf{0}$ and
$\mu_{12}=2c\mathbf{1}$ respectively. $F_2$ is assumed to be normal with identity covariance matrix
and $\mu_2=c\mathbf{1}$; we set $c=1$. This means that $F_1$ is symmetric bimodal and $F_2$ is
symmetric and lying between the two bumps of $F_1$. Figure \ref{FigExpNormMix} presents the results
of this configuration.

The flat performance of the LDA at 0.5 error rate is not a surprise; this is due to the fact that
the problem is symmetric and the hyper plane of symmetry, which will be the decision surface,
divides each distribution into two regions each has 0.5 probability. The QDA is the winner for its
ability to build quadratic surfaces capable of surrounding observations from $F_2$. In this
configuration the performance of the NCC gets worse as the dimensionality increases. This is in
contrast to the results of the first configuration. Moreover, at some training set sizes we get
$\operatorname*{E}_{\mathbf{tr}} Err > 0.5$, which means that the NCC rule has to be flipped to
produce a mean error rate of $1-\operatorname*{E}_{\mathbf{tr}} Err < 0.5$ . Therefore, the sign of
the rule varies with the training set size! and has to be determined by estimating the error rate
using one of the resampling techniques, e.g., cross validation. We can also notice that NCDA
outperforms NCC universally.

\bigskip

\begin{figure}[htb]
  \centering
  \subfloat[$p=2$]
  {\label{fig:exp12dmean}\includegraphics[width=0.4\textwidth]{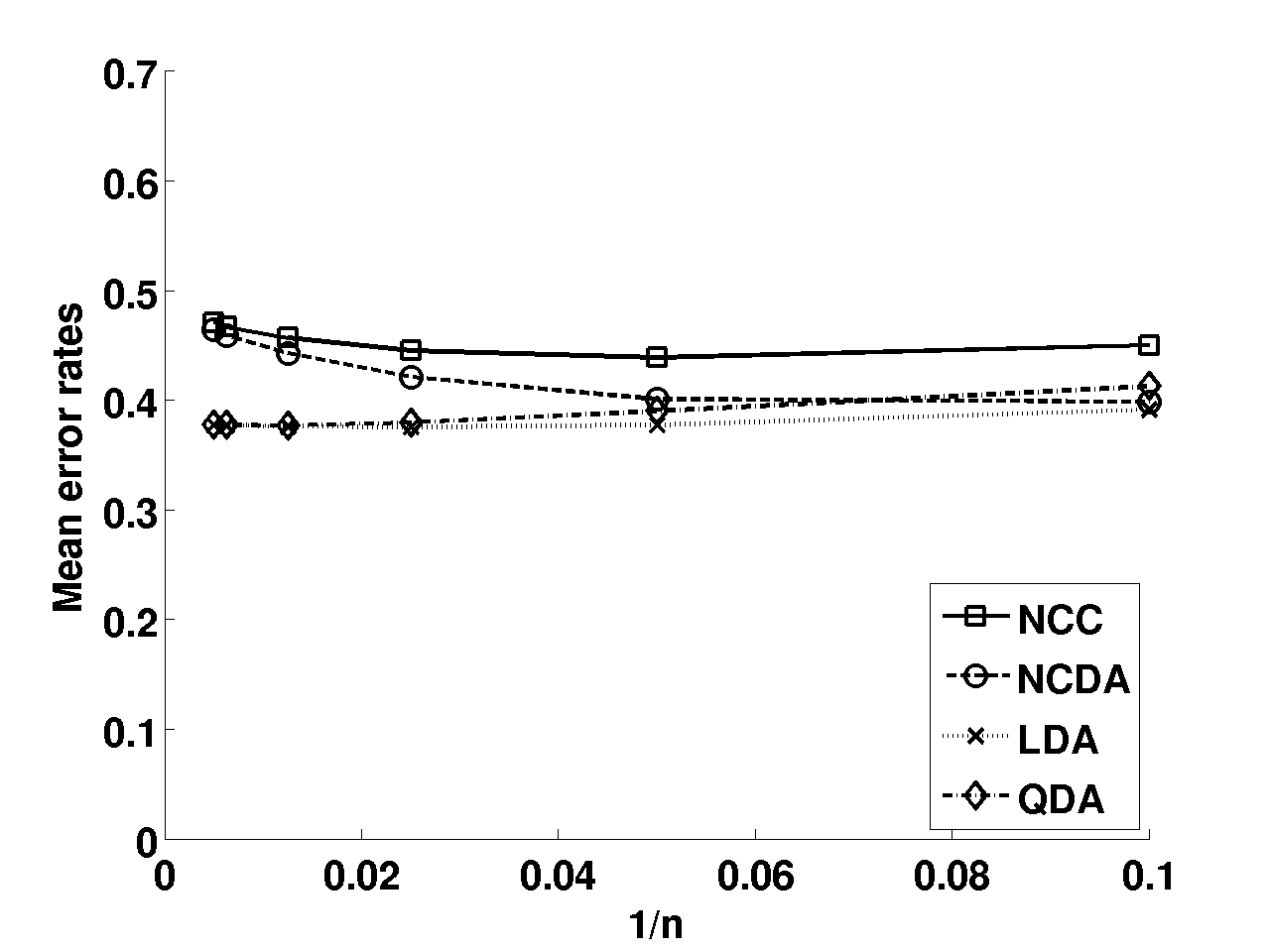}}
  \hspace{0.04\textwidth}
  \subfloat[$p=2$]
  {\label{fig:exp12dsigma}\includegraphics[width=0.4\textwidth]{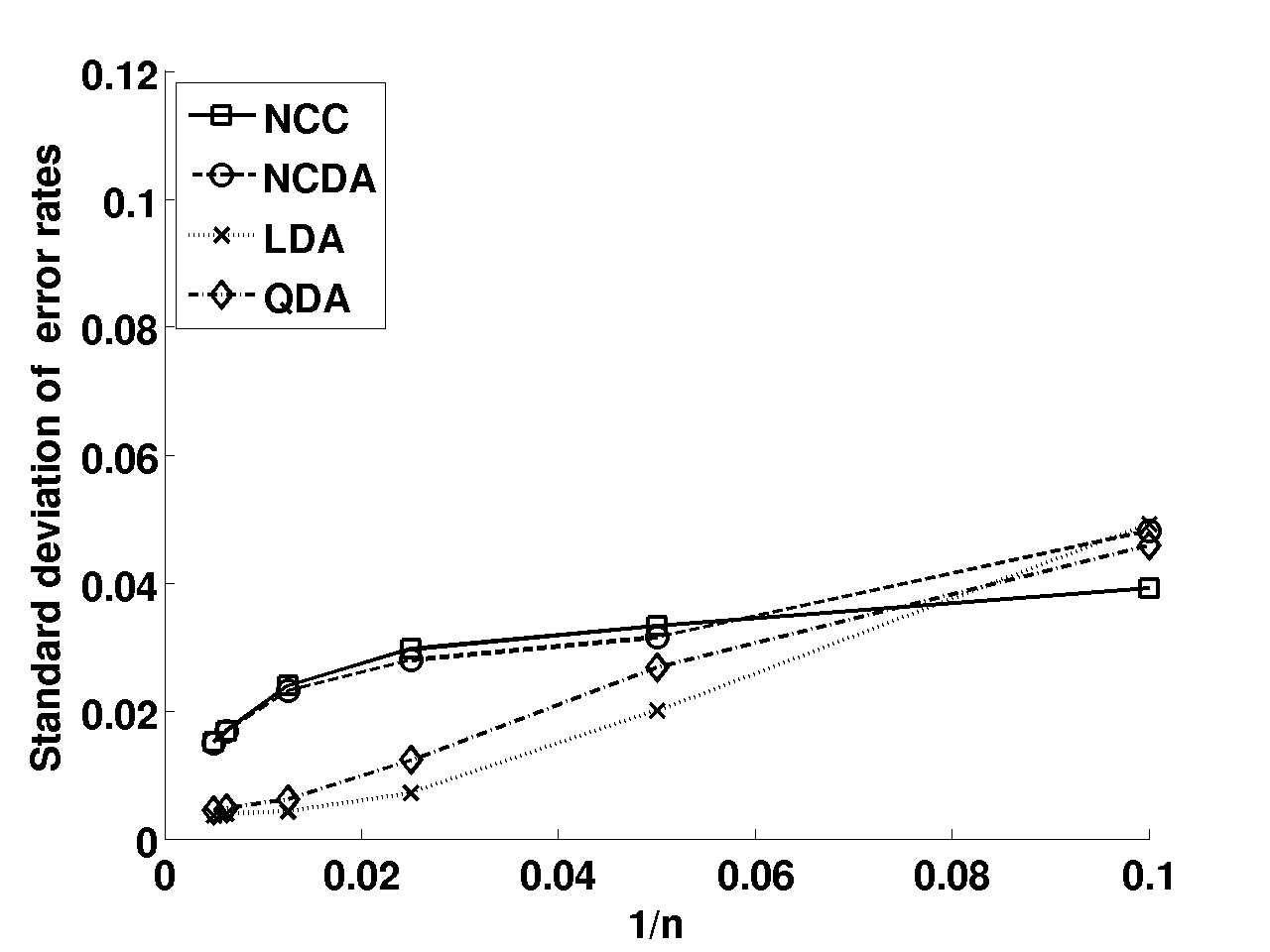}}
  \hspace{0.04\textwidth}
  \subfloat[$p=4$]
  {\label{fig:exp14dmean}\includegraphics[width=0.4\textwidth]{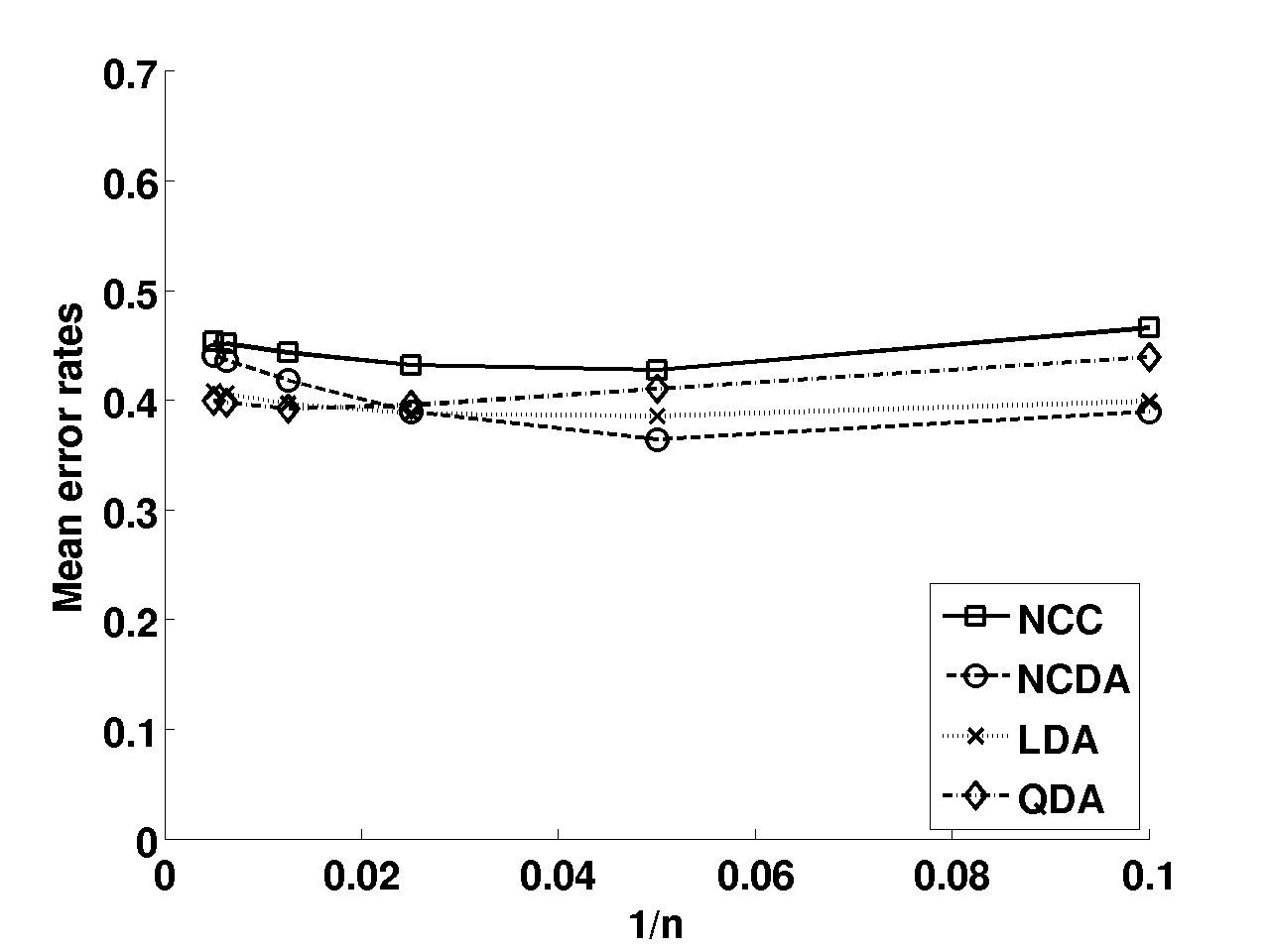}}
  \hspace{0.04\textwidth}
  \subfloat[$p=4$]
  {\label{fig:exp14dsigma}\includegraphics[width=0.4\textwidth]{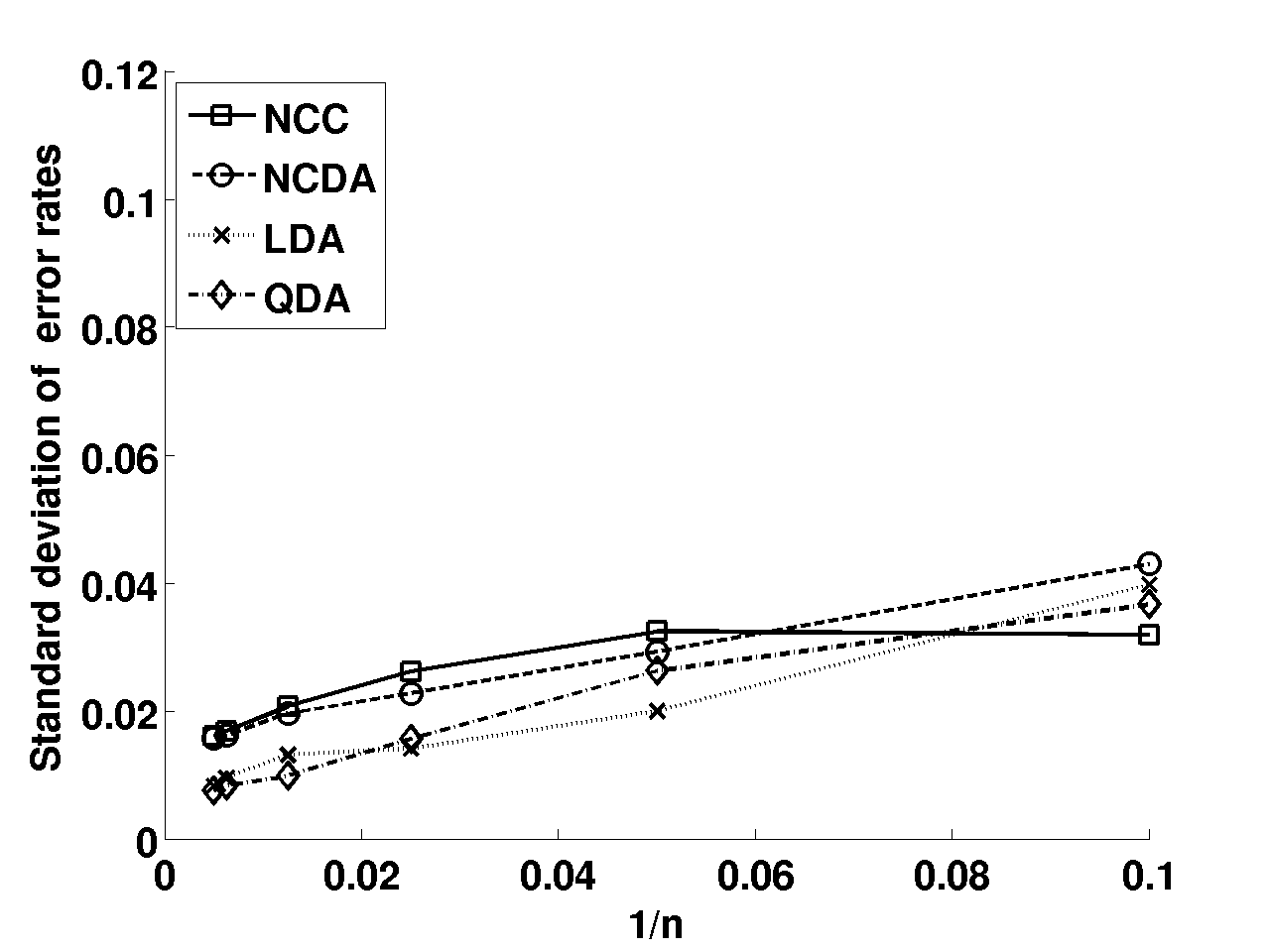}}
  \hspace{0.04\textwidth}
  \subfloat[$p=8$]
  {\label{fig:exp18dmean}\includegraphics[width=0.4\textwidth]{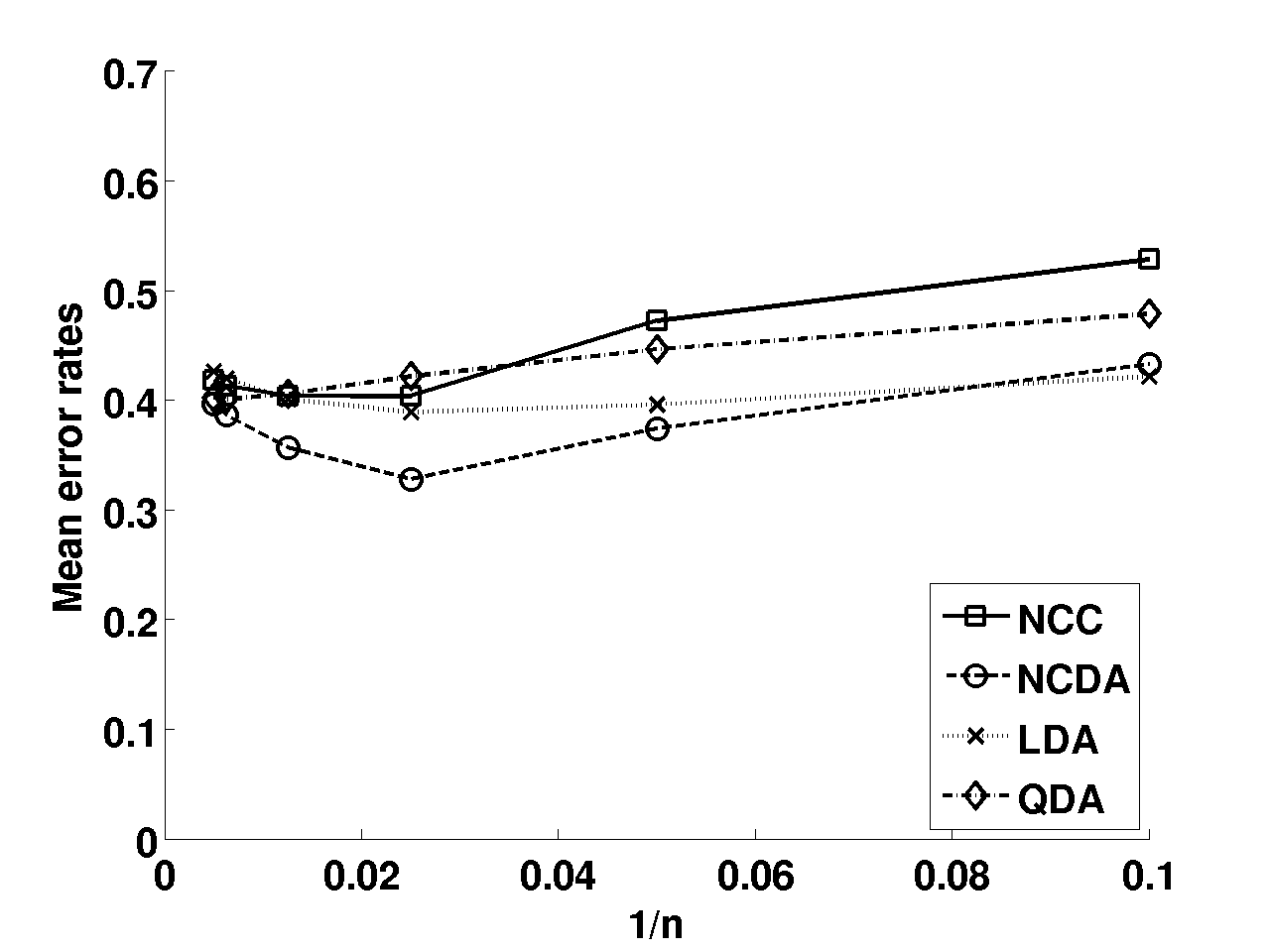}}
  \hspace{0.04\textwidth}
  \subfloat[$p=8$]
  {\label{fig:exp18dsigma}\includegraphics[width=0.4\textwidth]{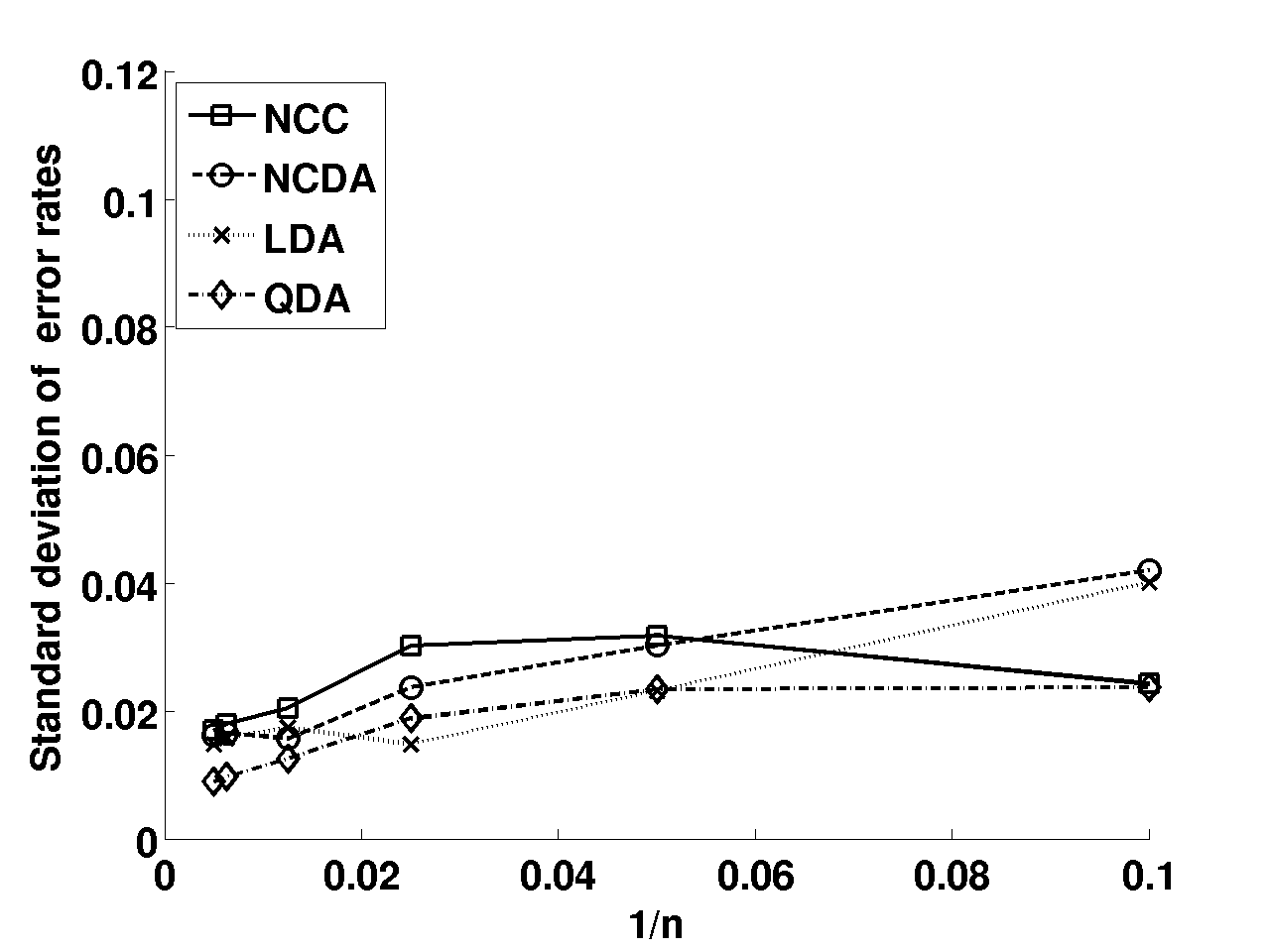}}
  \hspace{0.04\textwidth}
  \subfloat[$p=16$]
  {\label{fig:exp116dmu}\includegraphics[width=0.4\textwidth]{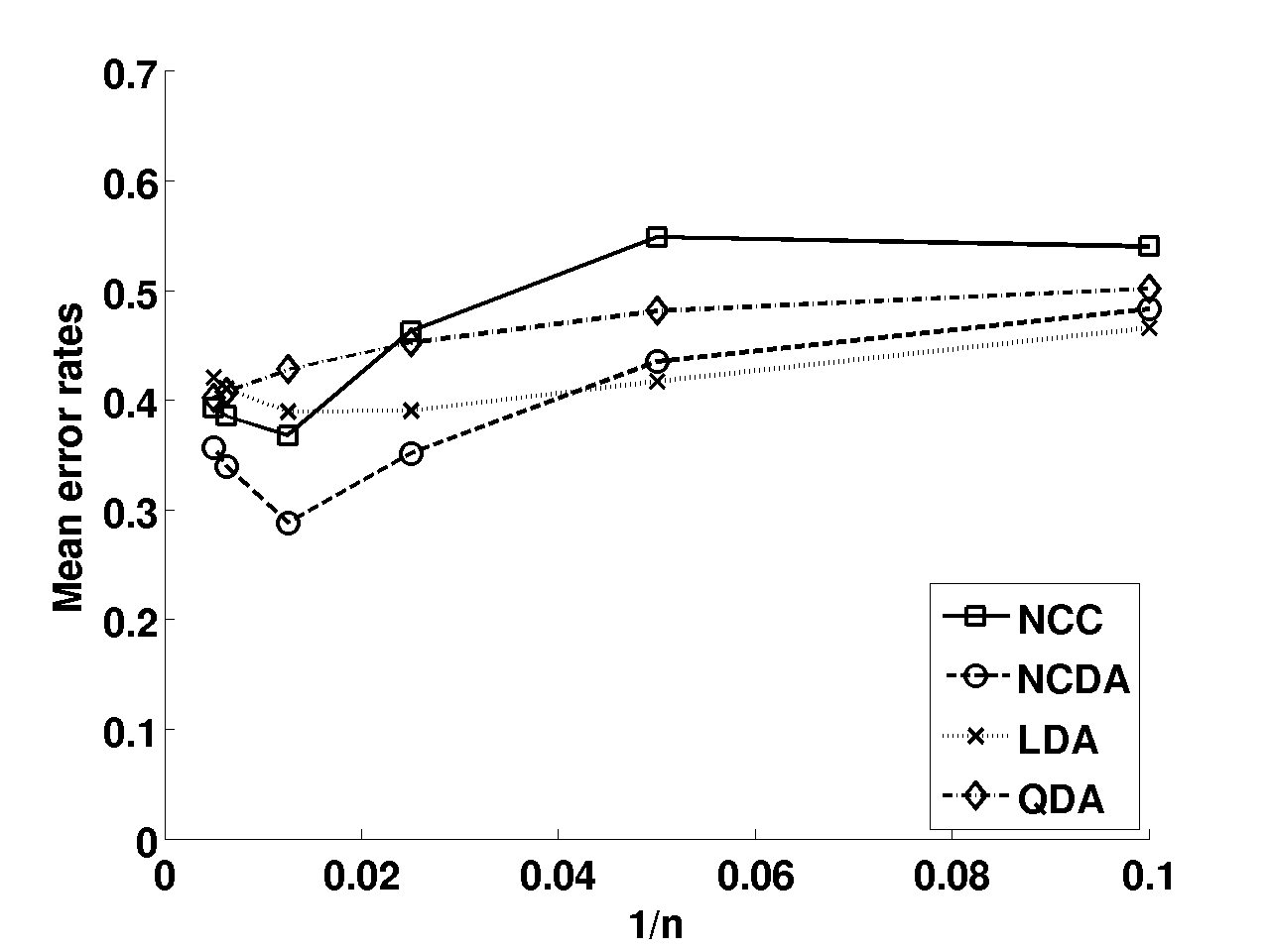}}
  \hspace{0.04\textwidth}
  \subfloat[$p=16$]
  {\label{fig:exp116dsigma}\includegraphics[width=0.4\textwidth]{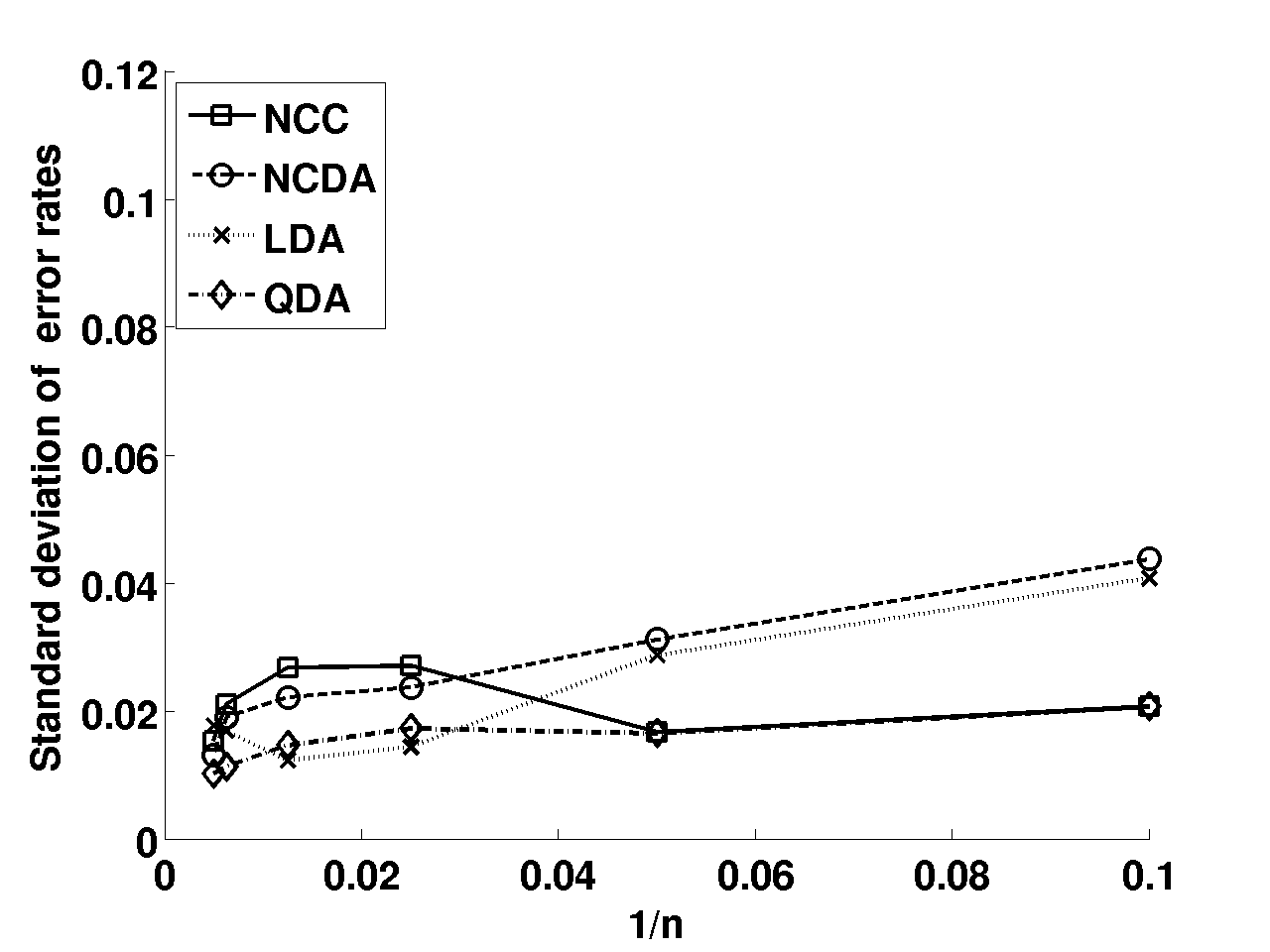}}
  \hspace{0.04\textwidth}
  \caption{Experiment 3 (mixture of two Gaussians, $\mu_1=\mu_2,\ \Sigma_1=\Sigma_2=I$): Mean (left) and standard deviation (right) of error rate under different $p$.}\label{FigExpMix}
\end{figure}

The third set of experiments assumes both $F_1$ and $F_2$ to be a mixture of two Gaussians (with two
different mean vectors and same identity covariance matrix). The first mean vector of $F_1$,
$\mu_{11}$, is set to zero vector. The second mean vector of $F_1$, $\mu_{12}$, is set to
$2c\mathbf{1}$. The first mean vector of $F_2$, $\mu_{21}$, is set to $c\mathbf{1}$. The second mean
vector of $F_2$, $\mu_{22}$, is set to $3c\mathbf{1}$; again, we chose $c=1$. Figure \ref{FigExpMix}
presents the results of this configurations.

From the figure we can observe that NCC, in the majority of experiments, is inferior to the other
three classifiers, while the NCDA outperforms all of them in many cases (except at $p=2$).

\section{Conclusion and future work}\label{sec:conclusion}
In this article we introduced a modification on the NCC classifier by combining it with the LDA; we
coined the name NCDA on the new classifier. We established a preliminary simulation study to compare
the performance of the NCC and NCDA to two basic classifiers, LDA and QDA. Our simulation study
reveals that the NCC is inferior to all other three classifiers almost at all considered dimensions,
training sample sizes, and distributions. This is in contrast to what has been reported in some
literatures \citep[e.g.,][]{Inselberg2002VisData}. Our proposed classifier, NCDA, outperforms NCC in
all experiments; moreover, it outperforms both LDA and QDA in some experiments.

\bigskip

Our future work, currently under progress in our group, considers several points. First, we are
planning for more comprehensive simulation study for more understanding of the behavior of NCC and
NCDA. Second, we always advocate for using the Area Under the receiver operating characteristic
Curve (AUC) \citep[see, e.g.,][]{Hanley1982TheMeaning} as a performance measure, since it is
independent of the threshold at which we make our decision. However, in the present article, we
measure the performance of a classifier in terms of the error rate for two reasons. (1) error rate
is the performance measure that was used in \cite{Inselberg2002VisData} to compare the NCC to other
classifiers. (2) measuring the performance in terms of the AUC only suits a classifier whose output
is given in terms of quantitative scores rather than binary decisions as NCC and NCDA. The work
currently under progress in our group is considering converting the NCC, and its smarter version
NCDA, to score-based classifiers to allow us to assess them in terms of the AUC
\citep{Yousef2019AUCSmoothness-arxiv, Yousef2019PrudenceWhenAssumingNormality-arxiv,
  Yousef2013PAUC}. In addition, we have the opportunity to apply a whole literature of nonparametric
estimation procedures including our methods: estimating uncertainty using influence function
\citep{Yousef2005EstimatingThe}, estimating uncertainty using UMVU estimation
\citep{Yousef2006AssessClass, Chen2012UncertEst, Chen2012ClassVar}, and estimating uncertainty using
cross validation estimators \citep{Yousef2019LeisurelyLookVersionsVariants-arxiv,
  Yousef2019EstimatingStandardErrorCross-arxiv}, among others.

\bibliographystyle{elsarticle-harv}
\bibliography{publications,booksIhave,booksINeed}

\end{document}